\documentclass{ecai} 

\usepackage{latexsym}
\usepackage{amssymb}
\usepackage{amsmath}
\usepackage{amsthm}
\usepackage{booktabs}
\usepackage{marvosym}
\usepackage{enumerate}
\usepackage{graphicx}
\usepackage{color}
\usepackage[dvipsnames]{xcolor}
\usepackage{algorithm}
\usepackage{algorithmic}
\usepackage{subcaption} 
\usepackage{newfloat}
\usepackage{fancyvrb}
\usepackage{mathtools}
\usepackage{threeparttable}
\usepackage{parcolumns}
\usepackage[normalem]{ulem}

\newcommand*{\img}[1]{%
    \raisebox{-.05cm}{%
        \includegraphics[
        height=.3cm,
        width=.3cm,
        keepaspectratio,
        ]{#1}%
    }%
}

\newcommand{\imply}{:\!\!-} 

\usepackage{newfloat}
\usepackage{listings}
\newfloat{listing}{tb}{lst}{}
\floatname{listing}{Listing}

\usepackage{pgf} 
\usepackage{colortbl}
\usepackage{etoolbox}
\definecolor{high}{HTML}{76f013}  
\definecolor{low}{HTML}{3377ff}   
\newcommand{\opacity}{50}         
\newcommand{\minval}{1.0}  
\newcommand{\maxval}{100.0}  
\newcommand{\gradient}[1]{
    \ifdimcomp{#1pt}{>}{\maxval pt}{#1}{
        \ifdimcomp{#1pt}{<}{\minval pt}{#1}{
            \pgfmathparse{int(round(100*(#1/(\maxval-\minval))-(\minval*(100/(\maxval-\minval)))))}
            \xdef\tempa{\pgfmathresult}
            \cellcolor{high!\tempa!low!\opacity} #1
    }}
}

\newcommand{\minvalb}{1}
\newcommand{\maxvalb}{500}
\newcommand{\ctime}[3]{
            \pgfmathparse{100-int(round(((log2((#1*36 + #2*0.6 + #3*0.01)+1)*100 -\minvalb)  / (\maxvalb - \minvalb ))*100))}
            \xdef\tempa{\pgfmathresult}
            \cellcolor{high!\tempa!low!\opacity}
            \ifdimcomp{#1 pt}{>}{0 pt}{#1h:}{}\ifdimcomp{#2 pt}{>}{0 pt}{#2m:}{\ifdimcomp{#1 pt}{>}{0 pt}{0m:}{}}#3s} 

\newcommand{\minvalD}{22}   
\newcommand{\maxvalD}{170}  
\newcommand{\ctimeD}[2]{
    \globalcolorstrue
    \pgfmathparse{100-(1+int(round(99*((#1*60+#2*1)/(\maxvalD-\minvalD))-(\minvalD*(99/(\maxvalD-\minvalD))))))}
    \xdef\tempa{\pgfmathresult}
    \cellcolor{high!\tempa!low!\opacity}
    \ifdimcomp{#1 pt}{>}{0 pt}{#1m:}{\ifdimcomp{#2 pt}{>}{0 pt}{0m:}{}}#2s
    \colorlet{lastcolor}{high!\tempa!low!\opacity}}

\newcommand{\minvalE}{92}  
\newcommand{\maxvalE}{411} 
\newcommand{\ctimeE}[2]{
    \globalcolorstrue
    \pgfmathparse{100-(1+int(round(99*((#1*60+#2*1)/(\maxvalE-\minvalE))-(\minvalE*(99/(\maxvalE-\minvalE))))))}
    \xdef\tempa{\pgfmathresult}
    \cellcolor{high!\tempa!low!\opacity}
    \ifdimcomp{#1 pt}{>}{0 pt}{#1m:}{\ifdimcomp{#2 pt}{>}{0 pt}{0m:}{}}#2s
    \colorlet{lastcolor}{high!\tempa!low!\opacity}}

\newcommand{\minvalF}{543}  
\newcommand{\maxvalF}{3681} 
\newcommand{\ctimeF}[3]{
    \globalcolorstrue
    \pgfmathparse{100-(1+int(round(99*((#1*60*60+#2*60+#3*1)/(\maxvalF-\minvalF))-(\minvalF*(99/(\maxvalF-\minvalF))))))}
    \xdef\tempa{\pgfmathresult}
    \cellcolor{high!\tempa!low!\opacity}
    \ifdimcomp{#1 pt}{>}{0 pt}{#1h:}{}\ifdimcomp{#2 pt}{>}{0 pt}{#2m:}{\ifdimcomp{#1 pt}{>}{0 pt}{0m:}{}}#3s} 

\newcommand{\minvalG}{24}  
\newcommand{\maxvalG}{441}  
\newcommand{\ctimeG}[2]{
    \globalcolorstrue
    \pgfmathparse{100-(1+int(round(99*((#1*60+#2*1)/(\maxvalG-\minvalG))-(\minvalG*(99/(\maxvalG-\minvalG))))))}
    \xdef\tempa{\pgfmathresult}
    \cellcolor{high!\tempa!low!\opacity}
    \ifdimcomp{#1 pt}{>}{0 pt}{#1m:}{\ifdimcomp{#2 pt}{>}{0 pt}{}{}}#2s
    \colorlet{lastcolor}{high!\tempa!low!\opacity}}

\definecolor{forest-red}{HTML}{F22C40}
\definecolor{forest-orange}{HTML}{DF5320}
\definecolor{forest-yellow}{HTML}{C38418}
\definecolor{forest-green}{HTML}{7B9726}
\definecolor{forest-cyan}{HTML}{00AD9C}
\definecolor{forest-blue}{HTML}{407EE7}
\definecolor{forest-violet}{HTML}{6666EA}
\definecolor{forest-magenta}{HTML}{C33FF3}

\definecolor{bright-blue}{HTML}{4477AA}
\definecolor{bright-cyan}{HTML}{66CCEE}
\definecolor{bright-green}{HTML}{228833}
\definecolor{bright-yellow}{HTML}{CCBB44}
\definecolor{bright-red}{HTML}{EE6677}
\definecolor{bright-purple}{HTML}{AA3377}
\definecolor{bright-grey}{HTML}{BBBBBB}

\colorlet{bright-blue-85}{bright-blue!85!white}
\colorlet{bright-cyan-85}{bright-cyan!85!white}
\colorlet{bright-green-85}{bright-green!85!white}
\colorlet{bright-yellow-85}{bright-yellow!85!white}
\colorlet{bright-red-85}{bright-red!85!white}
\colorlet{bright-purple-85}{bright-purple!85!white}
\colorlet{bright-grey-85}{bright-grey!85!white}

\definecolor{darkslategray3}{HTML}{79CDCD}
\definecolor{lightgoldenrod1}{HTML}{ffec8b}

\definecolor{wheat}{rgb}{0.96, 0.87, 0.7}

\newcommand{\BibTeX}{B\kern-.05em{\sc i\kern-.025em b}\kern-.08em\TeX}

\begin{document}


\begin{frontmatter}


\paperid{0737} 


\title{Answer Set Networks: GPU-Accelerated Answer Set Programming for Neurosymbolic AI}

\title{Neurosymbolic AI First! GPU-Driven Answer Set Programming} 

\title{Neurosymbolic AI First!\\ GPU-Accelerated Answer Set Programming}

\title{Answer Set Networks: GPU-Accelerated Answer Set Programming for Neurosymbolic AI }

\title{Answer Set Networks: GPU accelerated Deep Probabilistic Programming }

\title{ GPU accelerated Deep Probabilistic Programming\\ via Answer Set Networks }

\title{Answer Set Networks: A GPU accelerated solver for 
\\Deep Probabilistic Programming }

\title{Answer Set Networks: Mitigating the CPU bottleneck of Deep Probabilistic Programming }

\title{Answer Set Networks:\\ Leveraging GPU Acceleration At Every Step }

\title{Thinking Fast and Parallel:\\ Answer Set Networks for Rapid Neuro-Symbolic Systems}

\title{Thinking Fast and Parallel: Answer Set Networks\\ for Accelerated Neuro-Symbolic Systems}

\title{Thinking Fast and Parallel:\\ Accelerating Neuro-Symbolic AI by Answer Set Networks}
\title{Thinking, Fast and Not So Slowly:\\ Accelerating Neuro-Symbolic AI by Answer Set Networks}



\author[A]{\fnms{Arseny}~\snm{Skryagin}\thanks{Corresponding Author. Email: arseny.skragin@cs.tu-darmstadt.de}\footnote{Equal contribution.}}
\author[A]{\fnms{Daniel}~\snm{Ochs}\footnotemark}
\author[A]{\fnms{Philipp}~\snm{Deibert}}
\author[A]{\fnms{Simon}~\snm{Kohaut}}

\author[B]{\fnms{Devendra Singh}~\snm{Dhami}}
\author[A,C,D]{\fnms{Kristian}~\snm{Kersting}} 

\address[A]{AI \& Machine Learning Group, CS Dept., TU Darmstadt}
\address[B]{Uncertainty in AI Group, Dept.~of Mathematics and CS, TU Eindhoven}
\address[C]{Hessian Center for AI (hessian.AI)}
\address[D]{German Research Center for AI (DFKI)}

\begin{abstract}
Neural-symbolic (NeSy) systems such as Deep Probabilistic Programming Languages (DPPLs) have shown great potential by combining fast deep learning modules with the more deliberate symbolic reasoning, such as Answer Set Programming (ASP).
Although reasoning complements deep learning, it specifically suffers from the CPU-bound nature of even the most competitive solvers.
On top of this, NeSy inherits the requirement to process large datasets for training.
We propose Answer Set Networks (ASN), a GPU-based NeSy solver, to address both issues.
Based on Graph Neural Networks (GNN), ASNs offer an accelerated framework for ASP-based DPPLs.
Specifically, we show how to translate AS programs into ASNs and how ASNs can efficiently solve the encoded problem by leveraging a GPU's batching and parallelization capabilities.
Our overall contributions are twofold.
Namely, we are the first to show the fine-tuning of Large Language Models (LLM) with DPPLs, employing ASNs to guide the training with logic.
Furthermore, we scale the ``constitutional navigation'' of drones, i.e., encoding public aviation laws in an ASN for routing Unmanned Aerial Vehicles in uncertain environments. 
Our experimental evaluations demonstrate that ASNs outperform state-of-the-art CPU-bound NeSy systems on a selection of vision, language, and navigation tasks.
\end{abstract}

\end{frontmatter}


\section{Introduction}

NeSy systems combine high-level (probabilistic) reasoning with low-level perception modules. 
Among them, Deep Probabilistic Programming Languages~(DDPLs) offer a declarative way, to integrate various kinds of neural networks within logical reasoning. 
The first generation of DPPLs includes DeepProbLog~\citep{DeepProbLog}, followed by NeurASP~\citep{NeurASP}, SLASH~\citep{SLASH} and Scallop~\cite{Scallop}.
All of these languages share the same bottleneck: they are universally difficult to speed up because their reasoning relies on CPU-based solving.
Even if multithreading is possible, as shown in \citep{SLASH}, the speed-up is constrained due to logical queries being answered separately.
Hence, the anticipation for the second generation of DPPLs is to carry out both parts (reasoning and perception) on many core processors, to be able to unlock the potential of NeSy-AI.

Within the first generation, Scallop~\cite{Scallop} relies on \textit{top-k} to consider only parts of the solution space. 
Similarly, SAME~\citep{SAME} offers \textit{top-k\%} to dynamically prune unlikely solutions over time. 
While, both push the boundaries of DPPLs' scalability, the underlying problem persists, as queries are answered separately from one another.
Furthermore, in many NeSy applications, we make the observation that many samples share the same or similar queries. 
Consequentially, the number of unique queries is much smaller than the number of data entries. To the best of our knowledge, none of the existing DPPLs make use of this observation.
Another example in which the NeSy CPU bottlenecks become prevalent is Alpha Geometry~\cite{AlphaGeometry}. 
There, 10k CPU cores were used to train a model to reach the silver medal level in geometry at the math Olympiads.
Altogether, this suggests that the second generation should perform logical reasoning on the same hardware where deep model inference takes place; with solving routines designed with parallelization in mind to be easily vectorized to maximize the hardware utilization.



Modeling domain-specific problems in a logical language and then applying a general solving approach is a long-established paradigm.
To this end, Answer set programming (ASP), being a modern, non-monotonic reasoning framework based on stable model semantics, has experienced rising interest in NeSy inference systems. 
In this vein, recent NeSy AI pipelines~\cite{SAME,SLASH,NeurASP} have relied on a foundation of ASP for providing a declarative modeling language to encode rules and background knowledge. 
Although conflict-driven, state-of-the-art solvers like clingo~\cite{Potassco} and DLV~\cite{DLV} continually improve, today's ASP solvers are not immune to the data-intense world of the real-world NeSy applications.

Thus, the question of how to integrate these pipelines tightly with modern GPUs is more urgent than ever.
For example, large machine learning models for natural language and vision tasks have dramatically risen in popularity and public awareness.
Widely known, their success is greatly supported by their underlying architectures' ability to distribute work on many-core systems with ease.
Thus, translating the task of solving ASP-based modeling into a neural framework that leverages modern hardware acceleration to the same level can be seen as a key to applying NeSy systems in real-world applications. 


Usually, in state-of-the-art solvers, the search for the solutions, i.e., the stable models, of a program $\Pi$ among all interpretations follows a binary tree. 
That is, branches of the tree correspond to (partial) assignments of truth values to the program's atoms (partial interpretations).
Further, alternating between decision and propagation phases generates the searched stable models of the ASP. 
However, recent research has proposed alternative approaches based on the propagation of truth values within a graph.
Such graph-based Answer Set Programming solver systems~\cite{graph-asp} are based on \textit{Dependency Graphs} as a basis. 
The advantage of this explicit representation of atoms and literals as graph nodes is the uniform representation of stable models and preservation of causal relationships, meaning each atom is justified in a particular stable model.
Hence, the usual heuristics inherited from SAT solvers are avoided, i.e., the solving procedure is transparent.
\begin{figure*}[t] 
    \centering
    \resizebox{\textwidth}{!}{
        \includegraphics[]{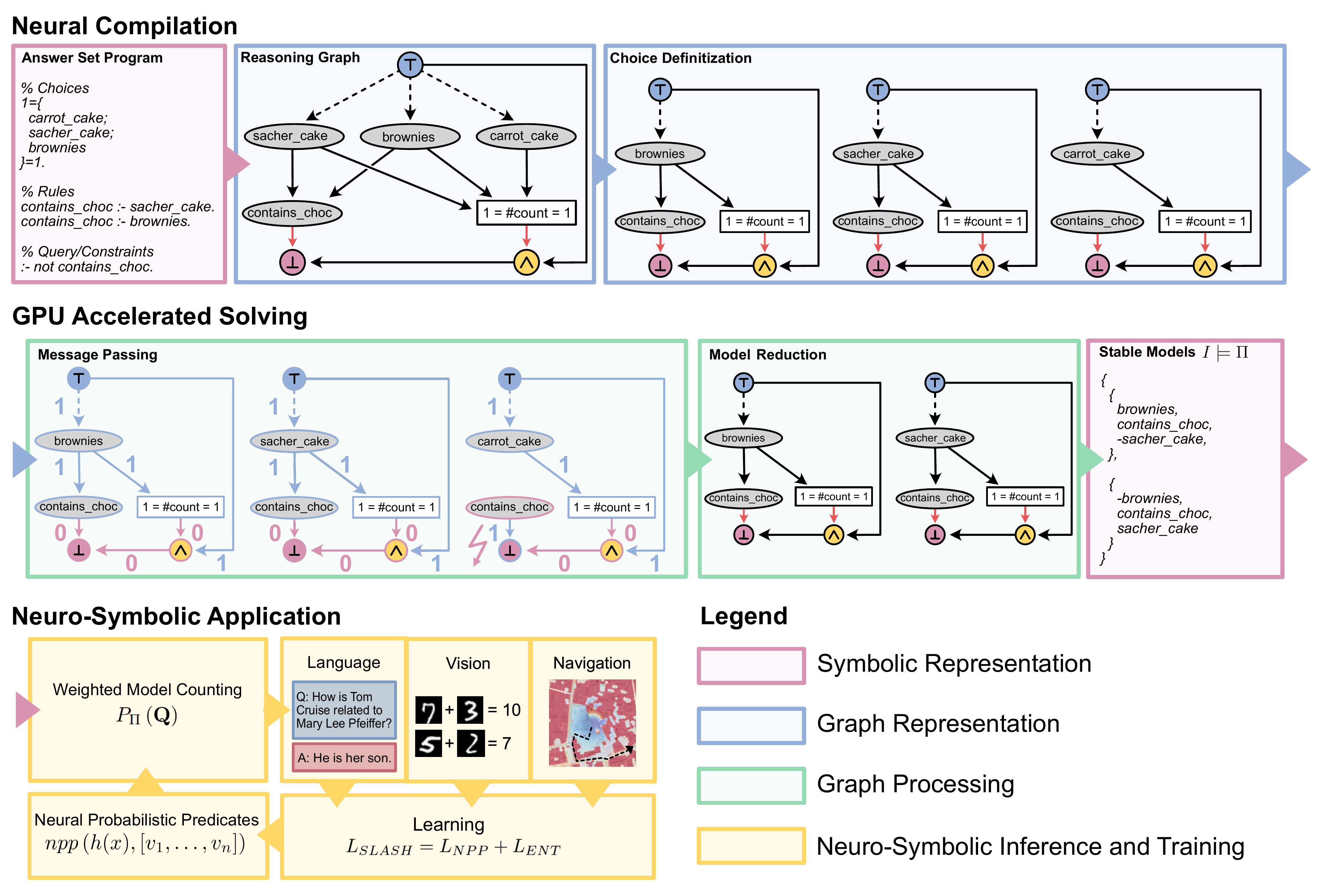}
    }
    \caption{
        \textbf{ASN from Answer Set Program to NeSy-AI Applications:}
        \textit{(left)} ASN takes a grounded ASP program as input and translates it into an equivalent Reasoning Graph via neural compilation.
        The RG instances representing all possible choice selections are constructed in the definitization stage, to be iteratively solved in parallel using message passing. 
        Finally, the resulting models are reduced to yield the ASP's stable models.
        \textit{(right)} Once stable models are in place, we can employ the Weighted Model Counting (WMC) to pursue the end-to-end learning for NeSy-AI applications in Natural Language Processing, Vision, and Navigation.
    }
    \label{fig:asn_inner_workings}
    \vspace{+.25cm}    
\end{figure*}

Along this line of research, we investigate a GNN-based representation of ASPs and the search for stable models.
More precisely, we present the novel NeSy ASP-solver \textit{Answer Set Networks} (ASN)\footnote{The plural signalizes that for solving, a stack of graphs is used.}\footnote{Code is part of supplementary material.}.\\ 
Within ASNs, we translate AS program into an equivalent Graph Neural Network (GNN) by introducing Reasoning Graphs (RG) as illustrated in Fig.~\ref{fig:asn_inner_workings}.
By introducing a set of building blocks as laid out in Fig.~\ref{fig:asp_to_gnn}, each element of an ASP program finds its respective representation within an RG.
ASN is motivated by grASP~\cite{graph-asp} to encode relationships and atoms as edges and nodes, respectively, to obtain stable models in a parallelizable way.

ASN as a NeSy solver of ASP problems, utilizes forward reasoning: 
First, they generate interpretations, i.e., labelings of the ground atoms.
Second, they filter through them to yield all the ASP's stable models.
ASN's motivation for the generation of interpretations is rooted in flow-networks.
Specifically, the ASN propagates truth values starting at a ``source'' node $\top$, from which the truth information can flow through the ASN until it reaches the `sink'' node $\bot$.
Hence, ASN can check whether an interpretation is valid or not according to the ASP. 

In summary, we make the following contributions:
\begin{itemize}
    \item We introduce Answer Set Networks, a novel ASP solver tailored towards scaling NeSy AI on many-core systems.
    \item We demonstrate the compilation of ASP programs into Reasoning Graphs and show how to obtain the respective stable models through message passing and model reduction. 
\end{itemize}
Our claims of ASNs outperforming state-of-the-art solutions in NeSy AI are substantiated in an exhaustive evaluation.
First, we demonstrate how ASN allows for fine-tuning of Large Language Models (LLMs) through probabilistic logic, alleviating the \textit{Reversal Curse}~\cite{reversalcurse}.
Second, ASN's raw performance benefits on the probabilistic mission design task for Unmanned Aerial Vehicles (UAV)~\cite{kohaut2023md}, showing three orders of magnitude faster inference than the baseline.
Finally, we investigate our contribution in detail in an ablation study showing empirical results on MNIST-Addition~\cite{DeepProbLog}. 

In the following, we introduce ASNs and how to obtain stable models from an answer set program. 
Precisely, we describe the process of neural compilation into an equivalent reasoning graph. 
Afterward, GPU accelerated truth value propagation via message passing to obtain the stable models is explained.
Before we go over the experiments to answer the research questions dedicated to demonstrating the advantages coming from ASNs.  
We give a brief overview of the related work and conclude with the outlook for future development.



\section{Interpreting Answer Set Programs as Graphs}
Here, we show how the ASP-Core-2~\cite{ASP-Core-2} language is encoded and solved in the form of the RG. 
ASN expects a grounded, i.e., variable-free and tight (acyclic) ASP-program, $\Pi$.
Fig.~\ref{fig:asn_inner_workings}~\textit{(left)} gives the overview of the ASN-pipeline, from a grounded ASP-program to \textit{stable models} (SM).
The pipeline consists of four steps, each of which we will describe in the following sections.
Before we continue with actual construction of the RG and the formal definitions, let us first build an intuition behind our approach to ASP solving. 
To summarize, ASN is inspired by \textit{control flow diagrams}.
I.e., each RG contains two special atom nodes, $\top$ (start) to be always true (its value is always 1) and $\bot$ (sink) to be always false (its value is 0 before message-passing).
To decide if a RG constitutes a candidate for SM, we control the value at $\bot$ after the final step of message-passing.
With other words, we differentiate two cases based on the final value of $\bot$: 
0 indicates a potential candidate and 1 means a contradiction, i.e., the sink cannot be true.
See~\Lightning~in Fig.~\ref{fig:asn_inner_workings}~(left, Message Passing) for example. 
The rest of this section and the one after is dedicated to formalization of every step by the example of choosing cake on the preference for chocolate (Fig.~\ref{fig:asn_inner_workings}~left).
Let us cake around and find out!

\subsection{Reasoning Graph}
\begin{figure*}[ht!]
    \centering
    \includegraphics[width=\textwidth]{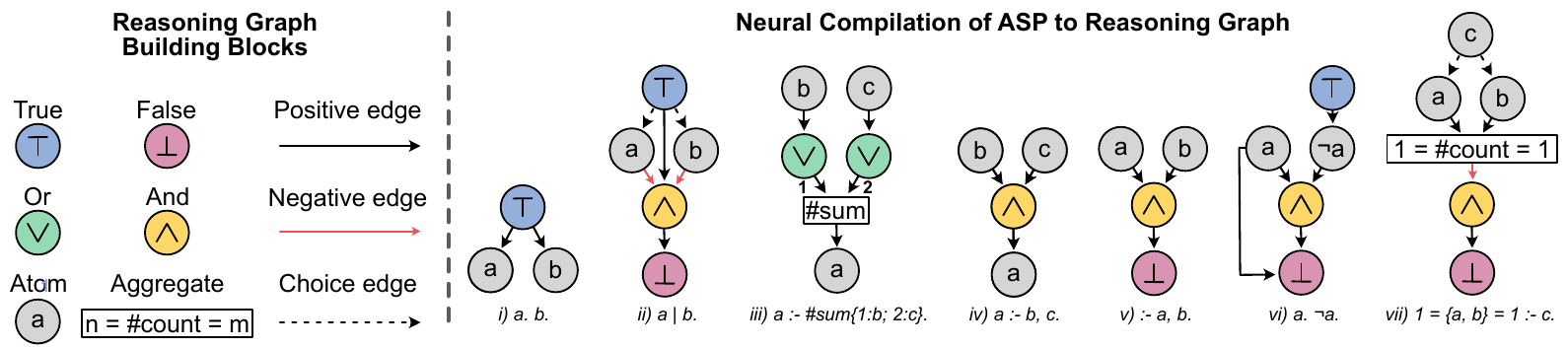}
    \caption{Buildings bocks of Reasoning Graphs and the RGs for the selection of ASP's syntax.}
    \label{fig:asp_to_gnn}
    \vspace{+.25cm}
\end{figure*}
We call the GNN representation of a grounded ASP-program $\Pi$ \textit{Reasoning Graph (RG)}, which we want to define formally as follows.
Let $\mathcal{G} = (\mathcal{V}, \mathcal{E})$ be a \textit{heterogeneous graph}, with a set of \textit{nodes} $\mathcal{V}$ and a set of \textit{edges} $\mathcal{E}\subseteq\mathcal{V}\times\mathcal{V}$.
Further, all nodes $v\in\mathcal{V}$ and edges $e\in\mathcal{E}$ have an associated \textit{type} $\tau_v\in\mathcal{A}$ and $\tau_e\in\mathcal{R}$, where $\mathcal{A}$ is the set of node types and $\mathcal{R}\subseteq\mathcal{A}\times\mathcal{A}$ is the set of edge types, respectively~\ref{fig:asp_to_gnn}. 
The tuple $(\mathcal{A},\mathcal{R})$ is called the \textit{network schema} of $\mathcal{G}$ and is itself a graph representing the node types and the possible relations between them.
A \textit{reasoning graph} is a heterogeneous graph where
\begin{align}
    \displaystyle\mathcal{A}&:=\underbrace{\{A_\land,A_\lor\}}_{\textstyle=:\mathcal{A}_\mathrm{logical}}\cup\underbrace{\{A_\mathrm{count},A_\mathrm{sum},A_\mathrm{min},A_\mathrm{max}\}}_{\textstyle=:\mathcal{A}_\mathrm{aggr}}\\
    \displaystyle\mathcal{R}&:=(\underbrace{\mathcal{A}\times\mathcal{A}_\mathrm{logical}}_{\textstyle=:\mathcal{R}_\mathrm{logical}})\cup(\underbrace{\mathcal{A}_\mathrm{logical}\times\mathcal{A}_\mathrm{aggr}}_{\textstyle=:\mathcal{R}_\mathrm{aggr}})
\end{align}%
The RG is constructed from a ground ASP program, where the nodes represent logical constructs -- namely classical atoms, aggregate literals, as well as conjunctions and disjunctions of the former two -- and the edges indicate rules for constructs to be true as defined by the semantics of the specified program $\Pi$. 

\subsection{Neural Compilation}\label{sec:neural-compilation}
Having the formal definition of the RG in place, the next step is to compile each element of the \textit{ASP syntax}.
I.e., the transition from ASP (red box) to RG (blue box) illustrated by Fig.~\ref{fig:asn_inner_workings}~(left).
We begin with the basic buildings blocks -- node and vertices types.
These are illustrated in Fig.~\ref{fig:asp_to_gnn}~(left).
Positive edges are shown in black, while red is used for negative ones.
Additionally, edges are annotated with their corresponding edge weights.
Dashed edges are parts of either disjunctive or choice-rules to indicate that they may or may not be active.
There are three categories in the ASP syntax: atoms, literals, and statements, which we will consider separately.

\subsubsection{Atoms}
As mentioned above, each RG has two special atom nodes, $\top$ (true) and $\bot$ (false), which, if connected to each other through other nodes, indicate true and false statements, respectively.
$\top$ is always part of any valid interpretation and propagates truth values. 
Facts, which are by design true, are always connected to the $\top$ node via an incoming edge (Fig.~\ref{fig:asp_to_gnn}~(right) \textbf{i})).
The $\bot$ node is used to indicate contradictions, model constraints, and ensures interpretation consistency.
Therefore, its purpose is to indicate the satisfiability of an interpretation $I$ or formally $I\models\Pi$.
We limit $\top$ to have no incoming edges and $\bot$ to have no outgoing edges, calling them \textit{source} and \textit{sink} nodes.
Apart from facts, other program atoms are defined through statements. 
For an atom to be true, at least one rule defining it must be satisfied.
In case there are multiple rules defining the atom, we express their combination through a disjunction.
Each unique classical atom is represented as an individual disjunction node. 
Note that predicate atoms and their negations are encoded separately. 

\subsubsection{Literals} 
Due to the program being already grounded, we know if a built-in literal evaluates to true or false. 
For conjunctions, true built-in literals can be ignored safely while making sure that false ones cannot be satisfied.
In the latter case, the conjunction can be treated as false as well.
For disjunctions, we change the course by ignoring false built-in literals and setting any to true if it evaluates to true itself.
Thus, during neural-compilation, built-in literals are handled implicitly.\\
\indent Next, we consider an aggregate literal of the form $\beta_{v,l} \diamond_{v,l}f\left\{T_i:C_i\right\}\diamond_{v,r} \beta_{v,r}$.
Depending on the problem to be encoded and the aggregate literal, the boundaries $\beta_{v,l/r} \diamond_{v,l/r}$ can be optional.
Herein, we consider \textit{aggregate function} $f\in\{\#\mathrm{count},\#\mathrm{sum},\#\mathrm{min},\#\mathrm{max}\}$, \textit{relational operators} $\diamond_{v,l/r}\in\{=,\neq,<,>,\leq,\geq\}$, \textit{terms} $\beta_{v,l/r}$, \textit{aggregate elements} $T_i:C_i$ with  $T_i$ and $C_i$ being sequences of terms and negation-as-failure (NAF) literals respectively.
See Fig.~\ref{fig:asp_to_gnn}~(right) \textbf{iii}) for an example of \#sum aggregate.
Each unique term $T_i$ is aggregated only once if satisfied, but may be satisfied by one or more NAF-literals $C_i$.
Let $T = \{T_1,\ldots,T_k\}$ be the set of terms appearing in the aggregate elements, and $C(T_i) = \{C_j|T_i:C_i\in\{T_1:C_1,\ldots,T_k:C_k\}\}$ the set of all conditions defining each $T_i\in T$.
A term $T_i\in T$ is aggregated if and only if some NAF-literal $C_j\in C(T_i)$ is satisfied. 
This can be rewritten as a disjunction of the different NAF-literals, $\bigvee_{C\in C(T_i)}C$.
It is then connected to an aggregate node $\#\mathrm{aggr}$ with edge weight $f(\{T_i\})$. 
Both, \textit{left-} $\beta_{v,l} \diamond_{v,l}$ and \textit{right-guard} $\diamond_{v,r} \beta_{v,r}$ are encoded in the aggregate node. 

\subsubsection{Statements}
\label{sec:statements}
A \textit{rule} has the general form $H \imply B.$ with the head, $H$ and the body, $B$.
Semantically, it is equivalent to a conjunction of the body literals.
This conjunction node combines the body literals via negative or positive edges, depending on whether the corresponding literal is default-negated or not.
For facts, the body consists of $\top$.\newline
\indent In compiling  $a\imply b_1,\ldots,b_n.$, the conjunction node is directly connected to the node representing the \textit{consequent atom} $a$, Fig.~\ref{fig:asp_to_gnn}~(right) \textbf{iv}).
\textit{Constraints}, $\imply b_1,\ldots,b_n.$, are treated the same way assuming the head atom $\bot$, signifying an invalid interpretation if the body is true, Fig.~\ref{fig:asp_to_gnn}~(right) \textbf{v}).\newline
For \textit{disjunctive rules}, $a_1|\ldots|a_m \imply b_1,\ldots,b_n.$, the conjunction node is connected to all consequent atoms $a_1,\ldots,a_m$ and each edge representing a choice to be made.
A subset of the edges may be removed in advance to represent a corresponding choice.
Further, at least one of the consequent atoms needs to be true if the body of the rule is true.
Consequentially, we add an appropriate constraint, 
\begin{equation}\label{eq:ens_conf}
    \imply b_1,\ldots,b_n,\mathrm{\mathbf{not }}\: a_1,\ldots, \mathrm{\mathbf{not }}\: a_m. 
\end{equation}
rendering the current configuration invalid if the body holds but none of the consequent atoms.
E.g., Fig.~\ref{fig:asp_to_gnn}~(right) \textbf{ii}) with an empty body.\newline
\indent The \textit{choice rule} is encoded directly:
Head atoms may have extra conditions that must be fulfilled in addition to the body. 
The conjunction node, representing the body, can, in this case, not be directly connected to the consequent atoms but needs to be combined with the disjunctions, representing that some condition for the individual head atoms is fulfilled.
\begin{figure}[t!]
    \centering
    \includegraphics[width=0.35\textwidth]{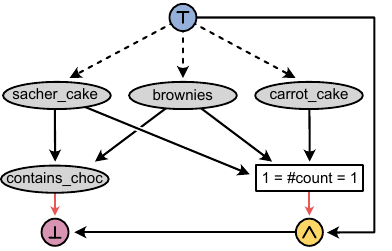}
    \caption{
        \textbf{The complete RG for the picking cake example}
    }
    \label{fig:cake_rg}
    \vspace{+.5cm}
\end{figure}
Additionally, it must be ensured that if the body is true, so are a valid number of head atoms as defined by the guards of the choice.
Thus, we add a similar to Eq.~\ref{eq:ens_conf} constraint.
\indent \textit{Weak constraints}, $:\sim b_1,\ldots,b_n.[w@l,t_1,\ldots,t_m]$ (with terms $w$, $l$, $t_1,\ldots,t_m$, $m>0$, $w$ for weight and $l$ for level), are ranking the resulting stable models and consequentially handled by the solver after computing all solutions without encoding into RG.\\
\indent \textit{Classical negation} can be regarded as syntactic sugar, where $\neg a$ is treated as a separate unique predicate symbol while keeping the same semantics.
To ensure consistency of a program $\Pi$, both $a$ and $\neg a$ cannot be part of the same model.
This requirement is encoded as the constraint  $\imply a, \neg a.$ with $\forall a,\neg a\in\Pi$, and is shown in Fig.~\ref{fig:asp_to_gnn}~(right) \textbf{vi}).

\indent \textit{Queries} in ASP are encoded as constraints and can be incorporated as such in the RG.
The key insight is that the majority of the program (up to the queries) is identical in all cases, and most of RG can be shared.
To separate the satisfiability values (i.e., the node values of the sink node) for different queries, we encode them using distinct sink nodes $\bot_{Q_1},\ldots\bot_{Q_n}$. 
The global sink node is then connected to these special sink nodes to pass along the satisfiability values of the main part of the program.

\section{GPU Accelerated Stable Models}
We now have seen the process of the neural compilation and saw how each ASP's syntax element is translated to be a part of the RG (following our running example in Fig.~\ref{fig:asp_to_gnn}~(left)).
The RG, Fig.~\ref{fig:cake_rg}, is represented by an equivalent GNN.
To make use of the idea of control flow diagrams for obtaining stable models, we ensure that the message passing leads to the expected behavior by making the RG \textit{definite}.
I.e., we replace all choice rules and disjunctions by their deterministic equivalents (Sec.~\ref{subsec:choice_definization}). 
After the choice definization, we proceed with message passing (Sec.~\ref{subsec:message-passing}), specifying how the values are calculated for each node type.
Finally, we read out all potential candidates to filter them for the minimal ones (Sec.~\ref{subsec:model_reduction}).
All of these steps are illustrated in green through the running example in Fig.~\ref{fig:asn_inner_workings}.

\subsection{Choice definitization}\label{subsec:choice_definization}
A RG may represent a program that contains disjunctive and choice rules. 
We consider an example program containing the single rule $a_1|a_2 \imply B.$ with $B$ being some sequence of NAF-literals.
This example can be represented by the set of programs $\{\{a_1\imply B.\},\{a_2\imply B.\},\{a_1\imply B.,a_2\imply B.\}\}$, which we will call \textit{program definitives} (or \textit{definitives} for short).
Each answer set of the original program consists of the \textit{program definitives}, making every distinct possible choice definite, transforming the original one into a normal program with a deterministic cause-effect relationship between classical atoms.
The RG representing these definitives can be directly constructed from a copy of the RG representing the original program, disabling (removing) the appropriate edges connecting the body to the non-selected consequent head atoms.
This covers the step from 
For programs containing multiple disjunctive, choice, and NPP-rules, all combinations of their respective variants need to be considered;
Which, in turn, yields easy parallelization of RG.

\subsection{Message-Passing}\label{subsec:message-passing}
Using message-passing, we want to find the final value at the sink node $\bot$ for each program definitive, as it is illustrated in Fig.~\ref{fig:asn_inner_workings} (left).
Thus, all atom nodes having value 1 (being true) in a candidate, the definitive not leading to a contradiction \Lightning, belong to a model.
Given a starting configuration of nodes that are considered true, we propagate truth values through the graph with the goal of computing the (stable) models of the program. 
Thus, we associate with every node $v\in V$ a boolean \textit{node value} $h_{v,t}\in\{0,1\}$ at a time step $t\in\mathbb{N}$. 
Except for $\top$, all nodes are initially set to false (0) at $t$ = 0.

Aggregate nodes $v\in\mathcal{V},\tau_v\in\mathcal{R}_\mathrm{aggr}$ additionally have terms $\beta_{v,l},\beta_{v,r}\in\mathbb{N}$ and relation operators $\diamond_{v,l},\diamond_{v,r}\in\{=,\neq,<,>,\le,\ge\}$ linked to them, representing the guards of the corresponding aggregate.
Furthermore, each edge $e\in\mathcal{E}$ has a \textit{weight}
\begin{equation}
    w_e\in
    \begin{cases}
        \{-1,1\}&\textrm{if }e\in\mathcal{E}_\mathrm{logical},\\
        \mathbb{Z} & \textrm{if }e\in\mathcal{E}_\mathrm{aggr}.
    \end{cases} 
\end{equation}
We can then define the node values for the next time step recursively as $\mathrm{h}_{v,t+1} =$
\begin{equation}
    \begin{cases}
        \mathrlap{\displaystyle \bigwedge_{\substack{e=(u,v)\in\mathcal{E},\\w_e=1}} h_{u,t} \land \bigwedge_{\substack{e=(u,v)\in\mathcal{E},\\w_e=-1}} \neg h_{u,t}} & \text{if } \tau_v=A_\land, \vspace{4pt}\\
        \mathrlap{\displaystyle \bigvee_{\substack{e=(u,v)\in\mathcal{E},\\w_e=1}} h_{u,t} \lor \bigvee_{\substack{e=(u,v)\in\mathcal{E},\\w_e=-1}} \neg h_{u,t}} & \text{if } \tau_v=A_\lor, \vspace{4pt}\\
        \beta_{v,l}\diamond_{v,l} |\mathcal{E}^{\mathrm{in},\top}_t(v)| \diamond_{v,r}\beta_{v,r} & \text{if } \tau_v=A_\mathrm{count}, \vspace{4pt}\\
        \beta_{v,l}\diamond_{v,l} \displaystyle\sum_{e\in\mathcal{E}^{\mathrm{in},\top}_t(v)} w(e) \diamond_{v,r}\beta_{v,r}  & \text{if } \tau_v=A_\mathrm{sum}, \vspace{4pt}
    \end{cases}
    \label{eq:rg_update_rule}
\end{equation}
where $\mathcal{E}^{\mathrm{in},\top}_t(v):=\{e|e=(u,v)\in\mathcal{E},h_{u,t}=1\}$ is the set of incoming edges of $v\in\mathcal{V}$ whose source nodes are 1 (true) at time step $t\in\mathbb{N}$ and $\mathcal{W}^{\mathrm{in},\top}_t(v):=\{w_e|e\in\mathcal{E}^{\mathrm{in},\top}_t(v)\}$ the set of corresponding edge weights.
We refer to App.~\ref{app:sec:formal_asn} for the definition of node values for aggregate functions of $\min$ and $\max$.
\\
\indent Updates at the time step $t+1$ can be understood in the following way. 
Logical nodes $v\in\mathcal{V}_\mathrm{logical}$ take on the value of the conjunction or disjunction of their incoming neighbors' values, where the edge weight determines whether the corresponding node value is included positively or negatively.
We therefore call an edge $e\in\mathcal{E}_\mathrm{logical}$ \emph{positive} if $w_e=1$ and \emph{negative} otherwise.
For aggregate nodes, $v\in\mathcal{V}_\mathrm{aggr}$, the edge weights of incoming edges signify the values to be aggregated, but are only included if the source node is considered true. 
The node value of an aggregate node is then determined by checking the guards of the aggregated value.

\subsection{Model Reduction}\label{subsec:model_reduction}
After message passing, the graphs represent interpretations, as indicated by the truth values of the nodes representing classical atoms. 
If $\bot$ is true for some graph instance, then the corresponding interpretation is not a model.
However, the interpretation does not equate to a stable model of the original program.
Accurately, ASP's models are minimal w.r.t. inclusion.
Thus, we filter all the program's models to obtain the minimal one.
This can be done efficiently in a batch-wise manner using logical bit-wise operators.
Let $I_1,I_2\subseteq A = \{a_1,\ldots,a_n\}\subseteq\mathcal{B}_{\Pi}$ be two interpretations over some finite subset of Herbrand base $\mathcal{B}_{\Pi}$. 
Further, let $m_1,m_2 \in \{ 0, 1 \}^n$ be vectors representing the truth values of the corresponding interpretations with the components $m_{i_j}=1$ if and only if $a_j \in I_i$.
We have $m_1 \otimes m_2 = $
\begin{equation}
    \begin{cases}
        m_1 & \text{if } I_1\subseteq I_2,\\
        m_2 & \text{if } I_1\supseteq I_2,\\
        m\in\{0,1\}^{n}, m_1\neq m\neq m_2 & \text{else}.
    \end{cases}
\end{equation}
where $\otimes$ represents element-wise logical conjunction (or multiplication). 
For a set of interpretations $\{I_1,\ldots,I_k\}\subseteq A$ with corresponding 0-1-vectors $m_1,\ldots,m_k$, the set of stable models can be computed as $ \mathrm{SM}\left(\Pi\right) = $
\begin{equation}
    \resizebox{0.9\columnwidth}{!}{$
       \left\{I_i|i\in\{1,\ldots,n\}: m_i\otimes m_j \neq m_i,\forall j = i+1,\ldots,n \right\}.
    $}
\end{equation}
This completes the last step of our example depicted in Fig.~\ref{fig:asp_to_gnn} (left), with us having two different cakes to pick from!

\section{Experiments}
This section has two objectives. First, we demonstrate how Neural Probabilistic Predicates (NPPs) can be encoded using ASN to integrate neural models within ASP for neuro-symbolic tasks. Then, building upon these NPPs, we show how ASN scales DPPLs with ASP as the backbone in various tasks by answering the following questions.
\begin{enumerate}
    \item[Q1] Which advantages offers \textbf{LLM fine-tuning with ASN} for the Reversal Curse~\cite{reversalcurse}?
    \item[Q2] Can ASN scale the task of \textbf{mission design for unmanned aerial vehicles}, as proposed by~\citep{kohaut2023md}, to cover Paris and its immediate surroundings?
    \item[Q3] How competitive can ASN be in a NeSy setting such as \textbf{MNIST-Addition}~\cite{DeepProbLog}? 
\end{enumerate}
Further, throughout the experiments, we refer to \textbf{ASN} as SLASH using ASN as ASP-solver and with \textbf{SLASH}~\cite{SLASH} the same DPPL but using clingo as solver.

\subsection{Neural Probabilistic Predicates}\label{sec:npp}
\begin{figure}[t]
    \centering
    \includegraphics[width=0.35\textwidth]{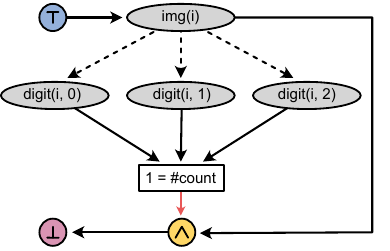}
    \caption{
        \textbf{Example of Neural-Probabilistic Predicate}: ASN encodes NPPs with  a choice rule. The choice atoms correspond to the outputs of a neural network, here a MNIST classifier with 3 digits.
    }
    \label{fig:npp_example}
    \vspace{+.5cm}
\end{figure}

To enable ASN for NeSy AI, we can encode \textit{Neural-Probabilistic Predicates} (NPP) from SLASH \cite{SLASH, SAME} within ASN:
\begin{equation}\label{eq:npp_rule}
    \#npp\left(h(x),\left[v_1,\ldots,v_n\right]\right) \leftarrow \textit{Body}
\end{equation}
Here, \textit{npp} is a reserved term labeling the NPP, 
\textit{h} a symbolic name of either a probabilistic circuit~\cite{ProbCirc20}, neural network or a combination of both. $v_1,\ldots,v_n$ denote the outcomes. As an example, lets consider the following NPP representing an MNIST classifier with three digits:
\begin{equation}
\#npp(digit(i), [0,1,2]):\!\!-\ img(i).
\end{equation}
In ASN we can represent NPPs by encoding them as choice rules, where each outcome of the neural network is a possible choice. Fig.~\ref{fig:npp_example} shows the corresponding graph. 
With NPPs, ASN allows for neural probabilistic programming and, simultaneously, probabilistic programming. 
For a thorough introduction to the SLASH semantics, we refer to~\cite{SAME}.
In general, SLASH can be seen as a superclass of ASP.

\begin{figure*}[t]
    \centering
    \resizebox{1.0\textwidth}{!}{
        \begin{subfigure}{.57\textwidth}
            \begin{subfigure}{.5\textwidth}
                \includegraphics[width=1.0\textwidth]{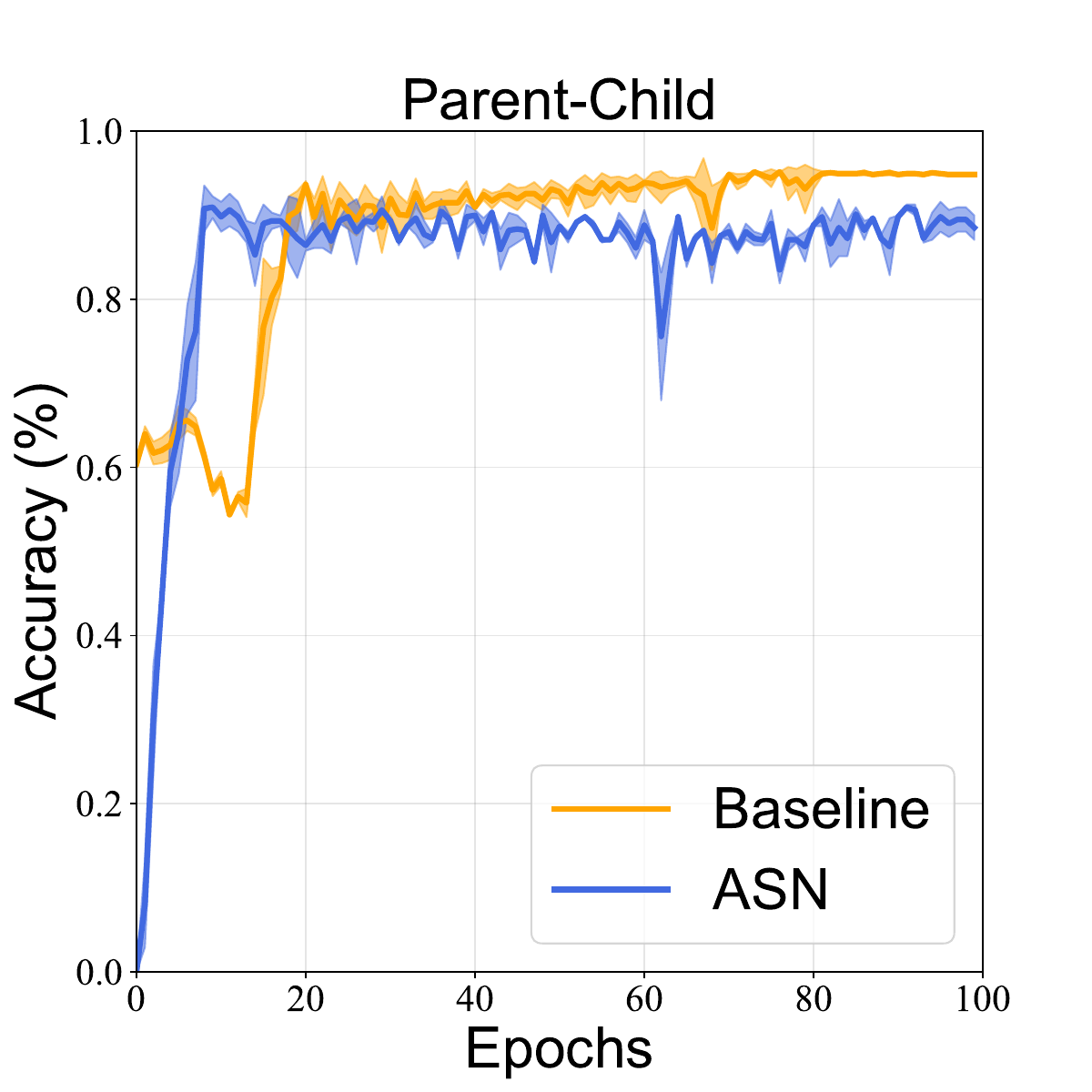}
            \end{subfigure}
            \begin{subfigure}{.5\textwidth}
                \includegraphics[width=1.0\textwidth]{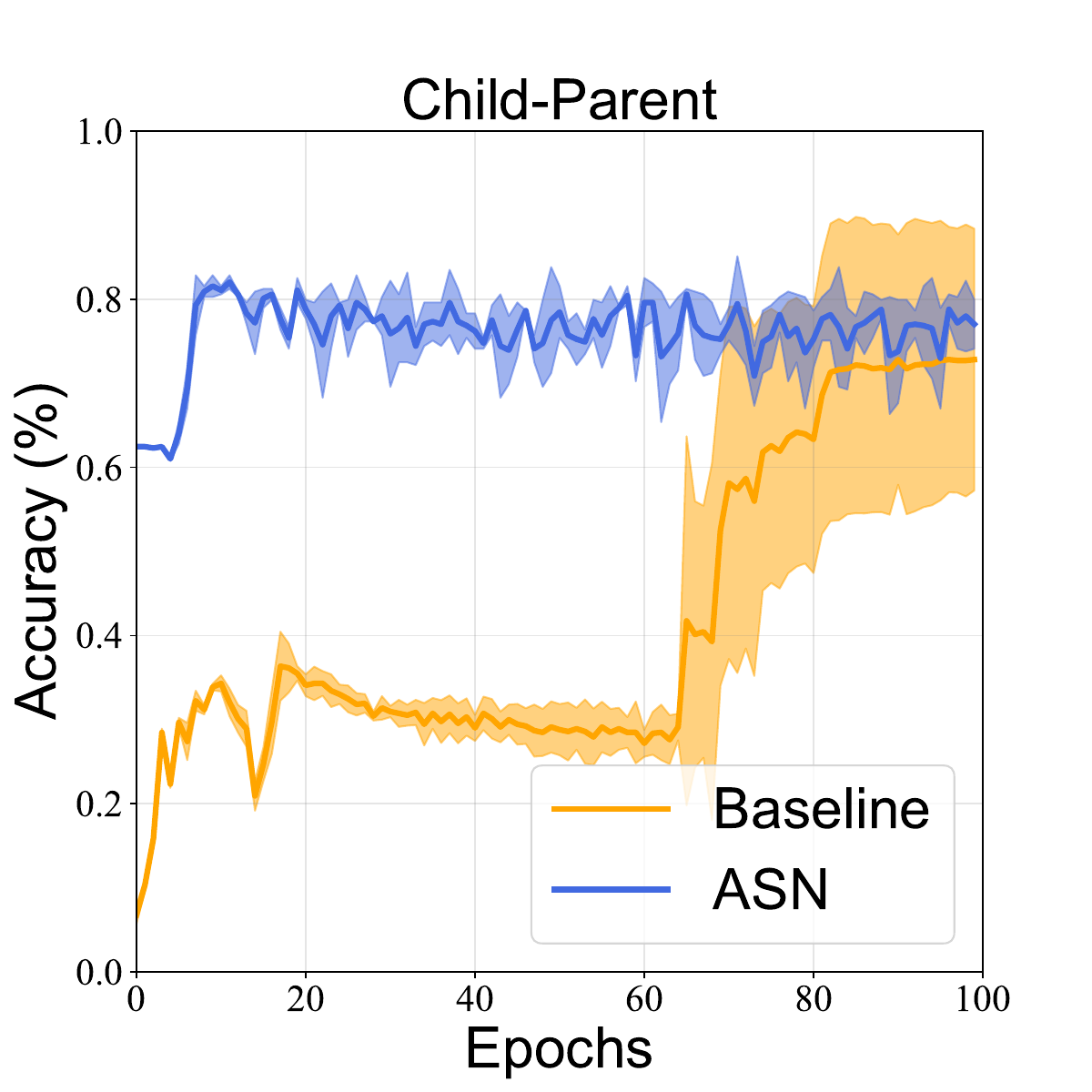}
            \end{subfigure}
            \caption{\textbf{ASN improves LLM's convergence during fine-tuning}: Learning to answer the query ``Q: Relation from Mary Lee Pfeiffer to Tom Cruise? A: Mother.'' (left) is easier than ``Q: Relation from Tom Cruise to Mary Lee Pfeiffer? A: Son.'' (right). Fine-tuning with ASN can significantly enhance convergence by supplying the LLM with the inverse signal, specifically $B\Rightarrow A$, derived from $A\Rightarrow B$, and vice versa.}
            \label{fig:abductive-fine-tuning}    
        \end{subfigure}
        \hspace{1.0mm}
        \begin{subfigure}{.27\textwidth}
            \centering
            \includegraphics[scale=0.215]{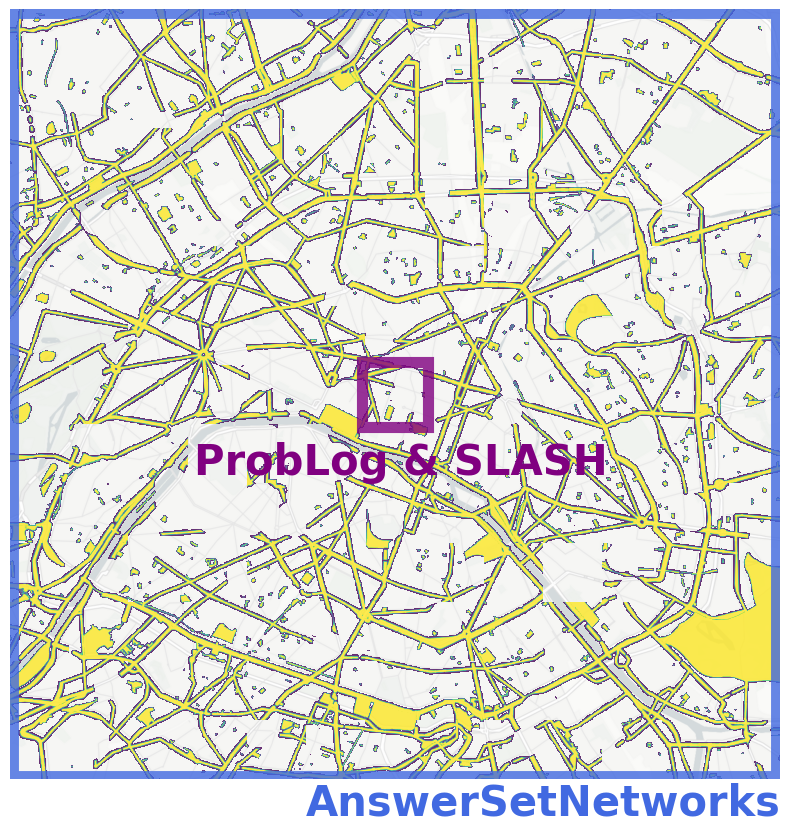}
            \caption{\textbf{ASN renders Paris in 56m}: Colored areas show where air travel is allowed, zero probability is uncolored. Meanwhile, SLASH \& ProbLog render a fraction of the map, shown by the purple square.
            }
            \label{fig:promis_paris}
        \end{subfigure}        
    } 
    \vspace{+.25cm} 
    \caption{ASN for LLM Fine-Tuning and ProMis over Paris during the Olympics.}
    \vspace{+.25cm}
\end{figure*}

\subsection[Q1]{Q1 LLM Fine-Tuning with ASN}
Berglund et al.~\citep{reversalcurse} have shown in their work that LLMs suffer from the Reversal Curse. 
Meaning, if a model is trained on a sentence of the form ``A $\Rightarrow$ B'', it will not automatically generalize to the reverse direction ``B $\Rightarrow$ A''. 
To this end, authors tested SOTA LLMs on pairs of questions like ``Who is Tom Cruise’s mother?'' and ``Who is Mary Lee Pfeiffer’s son?''.
Here, we demonstrate how ASN allows for fine-tuning LLMs to alleviate the problem.  

To address this issue, we perform \textit{fine-tuning via probabilistic logic} of the LLM using ASN.
Particularly, during training, we predict the relation (mother, father, daughter, and son) between a celebrity and their parent(s) in both directions.
The training dataset consists of queries such as ``Q: What is the relation between $<$celebrity\_name$>$ and $<$celebrity\_parent\_name$>$? A: ...'' and the opposite of the prompt by swapping $<$celebrity\_name$>$ and $<$celebrity\_parent\_name$>$. 
For each celebrity, we only provide one direction during training, i.e., either child-parent or parent-child.
We split the dataset into equal parts for each direction respectively to prevent the fine-tuned model from becoming biased in either direction. 
During test time, we evaluate the accuracy of unseen celebrities and their parents in both directions.

We add 12M learnable parameters by LoRA~\cite{LoRA} to fine-tune the Llama2-7B~\cite{llama2} LLM using ASN.
To compare the results of ASN fairly, we fine-tuned the Llama2 in a stand-alone fashion as well. 
The difference in the training procedure is founded in ASN's ability to derive the opposite relation, while baseline handles only one direction for each celebrity.
Using the original dataset from ~\cite{reversalcurse}, we split the 797 celebrities into
70\% train, 20\% test, and 10\% validation.
For in-depth explanations of the experimental setup, we refer our readers to consult the appendix.

The results reveal the following findings. 
Firstly, ASN's runtime equates to the number of passes through the LLM. 
Namely, we request both the sex and the relation for two persons for every query.
In particular, the baseline takes 50s and ASN 189s, which corresponds to the four forward passes and one backward pass. Importantly, this means we don't have any neurosymbolic overhead: The increased run-time is entirely due to the greater number of LLM passes which are connected with ASN.
Secondly, ASN converges one order of magnitude faster (in the number of epochs) than the baseline.
In detail, Fig.~\ref{fig:abductive-fine-tuning} (right) paints the picture for ASN taking only 9 epochs (28m:21s) vs. 96 (80m) for the baseline.
On the left, again, there were only 9 (28m:21s) for ASN and 72 (60m) for the baseline.
Lastly, ASN beats the baseline in the more complex direction ``B $\Rightarrow$ A'' (82.2\% vs 72.8\%) and performs slightly worse in the easier direction of ``A $\Rightarrow$ B'' (90.9\% vs 95.1\%).
Regarding \textbf{Q1}, these findings demonstrate that fine-tuning LLMs within a NeSy framework offers significant advantages while remaining computationally efficient. Moreover, this approach lays a strong foundation for future advances in guiding LLMs with probabilistic logic. 



\subsection[Q2]{Q2 Mission Design for Unmanned Aerial Vehicles}
    
    
    
    
    
Logic is crucial to ensure safety constraints, e.g., in autonomous navigation.
This is evident when considering autonomous agents navigating in human-inhabited spaces.
For example, a complex set of rules must be enforced when considering novel applications of Advanced Aerial Mobility, such as in cutting-edge logistics tasks. 
Similarly, metropolitan areas and capital cities pose a particular challenge due to their complex infrastructure and large scale. 
Within their borders, they often contain consulates of numerous countries, government buildings, etc.
They can also host major events that attract numerous international guests from all over the world.
All of this makes navigation particularly challenging because such events require additional temporary regulations that must be adhered to.
For example, Paris, France's capital, was the host city for the 2024~Olympic Games.
Hence, in this experiment, we show how ASN scales NeSy methods for safe navigation to such large-scale scenarios by densely sampling Paris' area. 

In the ProMis framework~\cite{kohaut2023md}, the requirements of public laws and operator preferences are encoded in a probabilistic logic language.
Essentially, one obtains i.i.d. generative models for each continuous point in the agent's navigation space.
Although one can compute probabilities independently, splitting up work and parameters in a multithreaded setting, a large number of logic programs need to be solved.
For each $(x_i, y_i) \in \mathcal{S}$ of points $\mathcal{S}$ in a two-dimensional navigation space, we query for a valid \textit{airspace}$(x_i, y_i)$.
All models entailed via the constraint that encodes the query fulfill one of the rules defining wherever the UAV is allowed to visit the specified point.


ProMis obtains a probability for each coordinate $(x_i, y_i)$ via Hybrid ProbLog~\cite{HybridProbLog}.
Using a stochastic error model~\cite{flade2021error}, it converts crowdsourced geographic features from OpenStreetMap into probabilistic facts.
These probabilistic facts encode probabilities for spatial relations between coordinates and tagged features, e.g., whether a coordinate is likely over a public park or far away from the nearest sports arena.
Note that the probabilities are ``hard-coded'' as they are obtained from the input map and not forwarded through a neural network, i.e., we use NPPs to encode the probabilities within the ASP program and use SLASH's inference mechanism without training.
Via SLASH, we can then compute the query probability $p(Q)$ from the stable models, which ASN computes for each coordinate.

To cover the area of 169km\textsuperscript{2}, we must solve 6500\textsuperscript{2} programs corresponding to all coordinate query pairs.
The resulting map can be seen in Fig.~\ref{fig:promis_paris}.
Despite the problem's size and complexity, ASN successfully generated the visible map in \textbf{56m:32s}.
We took the time to compute a 500\textsuperscript{2} grid for ProbLog and SLASH and multiplied it by 169 to obtain the time it would take to compute the whole grid. 
For ProbLog, it would take \textbf{7d:15h} and \textbf{5d:9h} for SLASH.
This makes ASN 194 and 137 times faster on ProMis than ProbLog and SLASH, respectively.
These findings answer \textbf{Q2} strongly, and we conclude that ASN's scalability opens up the opportunity to investigate further related problems. E.g., such as automated route planning~\cite{stenger2013route} and reactive navigation~\cite{Patrinopoulou23}. 
For further details on hyperparameter tuning of batch-size and other details, we refer to the appendices. 

\subsection[Q3]{Q3 MNIST-Addition}
In the MNIST-addition task~\cite{DeepProbLog}, the objective is to predict the sum of two images from the MNIST dataset, where the images are provided as raw data. 
During testing, the model must classify the images directly without explicit information about the digits they depict. 
Instead, it learns to recognize the digits through indirect feedback on the sum prediction. 
Handling more than two images significantly increases the task complexity due to the exponential growth in possible digit combinations. 
Similar to SAME~\cite{SAME}, we assess the model's scalability across three difficulty levels. 
These levels range from task T1, depicting two images with a target sum, e.g. $\mbox{\it sum2}$(\img{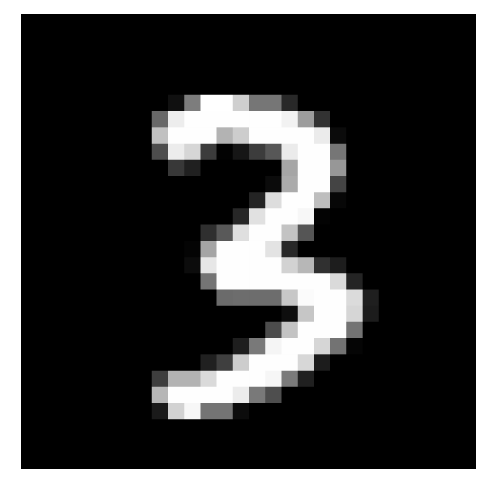},\img{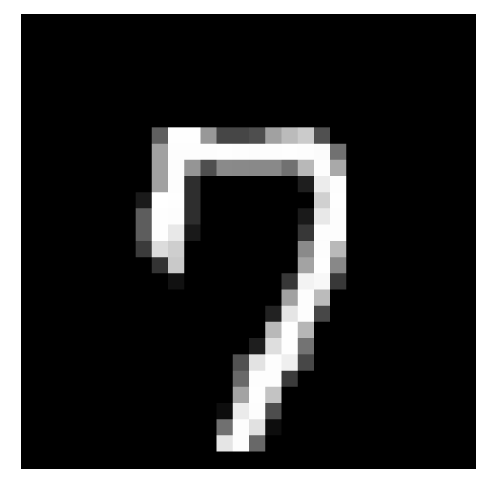},10), to task T3, which includes four images depicting digits as in $\mbox{\it sum4}$(\img{images/handwritten_digit_three.pdf},\img{images/handwritten_digit_seven.pdf},\img{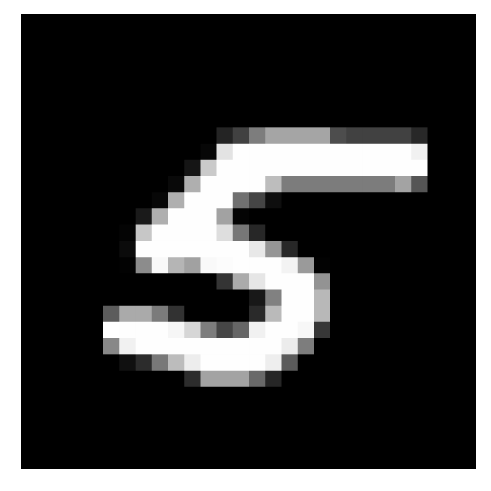},\img{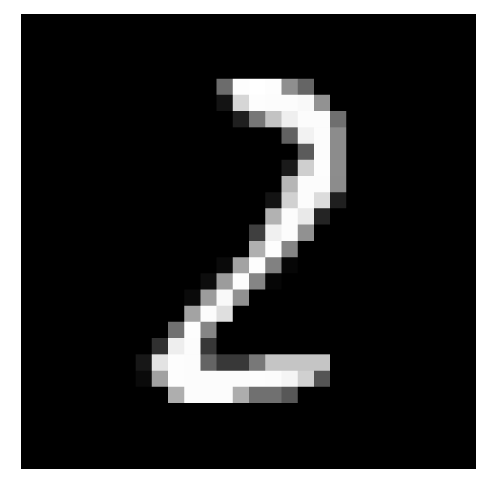},17). 
We evaluate performance using LeNet5~\citep{LeNet} as a NPP for the same experimental settings.
\begin{table}[t!]
    \centering
    \resizebox{1.0\columnwidth}{!}{
        \begin{tabular}{lrrr|rrr}
        \toprule
         & \multicolumn{3}{c}{Accuracy after last Epoch} & \multicolumn{3}{c}{Average Time per Epoch} \\
         Method & \multicolumn{1}{c}{T1}&\multicolumn{1}{c}{T2}&\multicolumn{1}{c}{T3} &\multicolumn{1}{c}{T1}& \multicolumn{1}{c}{T2}& \multicolumn{1}{c}{T3}\\
         \midrule
         DeepProbLog &\gradient{98.50}&\gradient{98.75}&\gradient{98.23} &\ctime{0}{8}{3}&\ctime{0}{15}{36}&\ctime{0}{34}{54}\\
         SLASH &\gradient{98.80}&\gradient{98.85}&\gradient{98.75} & \ctime{0}{0}{24}&\ctime{0}{1}{42}&\ctime{0}{51}{49}\\
         SAME &\gradient{98.56}&\gradient{98.82}&\gradient{98.71} & \ctime{0}{0}{17}&\ctime{0}{0}{17}&\ctime{0}{1}{35}\\
         \midrule
         ASN &\gradient{98.83}&\gradient{98.73}&\gradient{98.47} &\ctime{0}{0}{5}&\ctime{0}{0}{9}&\ctime{0}{0}{35}\\
         \bottomrule    
        \end{tabular}
    }
    \caption{\textbf{ASN scales well with growing task complexity:} Test accuracy in \% and runtime comparison for MNIST-Addition task. The runtime is averaged over ten epochs and five seeds for all methods. Light green indicates high accuracy or low time, while blue represents the opposite.
    }
    \vspace{+.25cm}
    \label{tab:mnist-comparison}
\end{table}

Following the results in Tab.~\ref{tab:mnist-comparison}, we observe that ASN is a highly beneficial solver for SLASH~\cite{SLASH}. 
Regardless of the tasks' complexity, ASN provides the big leap in computational speed, surpassing SAME~\cite{SAME}: T1 is 3,4 times faster, T2 - 1,89, and T3 - 2.71. 
These results answer \textbf{Q3} positively and confirm that seamlessly integrating subsymbolic models for NeSy tasks via ASN should be investigated further.
We refer to appendix for an in-depth discussion about choices of batch size, the example of a program, and its RG.


In summary, we have demonstrated in the experimental section how ASN positively impacts NeSy tasks that have to be solved multiple times with different configurations.
For these problems, which are at the heart of NeSy AI, the proposed ASN bridge the gap by bringing the symbolic part to the neural part, which lies on the GPU.

\section{Related Work}
\subsection{Automated Reasoning}
Logic programming traditionally relies on algorithms like Davis-Putnam-Logemann-Loveland (DPLL) for backtracking search~\citep{DPPLalgorithm}, augmented with no-good learning~\cite{no-good} and conflict-driven clause learning (CDCL)~\cite{CDCL}. 
One can make the distinction of backward and forward reasoners which start by the query and work backwards to the facts or the other way round. ASP implementations like clingo and DLV exemplify the latter approach, first grounding programs to propositional form then applying CDCL-based solving~\cite{CDCL}. Satisfiability Modulo Theories (SMT) extends these foundations by integrating domain-specific theories (e.g., arithmetic, data structures) with Boolean satisfiability checking~\cite{improvement-of-SAT-and-SMT-Solvers}.

Although the field of SAT and SMT solvers is competitive and consistently improves at solving complex problems~\cite{improvement-of-SAT-and-SMT-Solvers}, single-core execution inherently limits their potential~\cite{single-core-limitations} for NeSy applications.
Though some multithreaded extensions exist~\cite{clasp2}, few leverage many-core parallelism effectively. ASN addresses this gap by introducing a parallel-by-design solving mechanism based on GNNs.


\subsection{Probabilistic Logic Programming}
Probabilistic inference over a model's potential solutions is necessary to capture ambiguities in the modeled problem.
This is the case in Visual Question Answering problems~\cite{vqaintroduction}, with various datasets and benchmarks emerging~\cite{ok-vqa,a-okvqa,clevr,sort-clevr}.
Probabilistic logic languages, like ProbLog~\cite{ProbLog}, propose Weighted~\cite{introduction_to_WMC} and Algebraic Model Counting~\cite{AMC} over a foundation of SAT-based Prolog.
Many approaches have emerged to accelerate the model counting (inference) step, e.g., through structural adaptations or by porting computations to GPU's~\cite{KLay} in a similar vein to our approach.
As these systems depend on automated reasoning, as discussed above, the performance gains of ASN directly translate to probabilistic logic languages.

\subsection{Neural-Symbolic Systems}
The NeSy approach combines neural and symbolic methods to leverage their complementary strengths. Systems like DeepProbLog~\cite{DeepProbLog} and Scallop~\cite{Scallop} extend the semantics of ProbLog with neural predicates, enabling the integration of neural perception (e.g., image recognition) into probabilistic logic.
Similarly, SLASH~\cite{SLASH} and NeurASP~\cite{NeurASP} are DPPLs build on ASP.
However, all of these approaches are often limited by the high computational cost of symbolic reasoning.
Inspired by Scallop, Dolphin~\cite{dolphin2025} and Lobster~\cite{lobster2025} adopt a Datalog-based approach and accelerate inference on GPUs using Differentiable Top-k Proofs, Add-Mult Probabilities, and other differentiable provenance semirings.
KLay~\cite{KLay} complements this direction by organizing arithmetic circuits into knowledge layers, optimized for fast parallel execution on GPUs.
Currently, no ASP-based DPPL supports solving on many-core hardware. By integrating ASN into SLASH, we present the first such system and demonstrate its scalability for neurosymbolic tasks.


\section{Conclusions and Future Work}
We introduced Answer Set Networks, a novel NeSy solver for Answer Set Programs using Reasoning Graphs. ASN integrates seamlessly with ASP-based Deep Probabilistic Programming Languages like SLASH. Our experiments demonstrate its effectiveness across various NeSy tasks. Notably, to the best of our knowledge, we are the first to fine-tune large language models within a DPPL framework.

Through Answer Set Networks, we have taken important steps toward unlocking a wide range of future NeSy AI applications by leveraging the scalability of GPU-based computations.
Hereby, ASNs allow for distributed graph learning, allowing the symbolic component to be scaled alongside the neural component to larger problems and datasets.
We acknowledge that symbolic reasoning is still NP-hard and observe that when the number of NPPs increases we must deal with an exponential number of choice definitizations. 
To deal with even more complex problems, the next step is to integrate principles from CDCL into the message passing routine to eliminate unpromising choice definitizations and, hence, reduce the search space for potential solutions. 
Additionally, leveraging GNNs as the backbone of RG opens the door to developing differentiable weighted-model semantics for ASN.




\begin{ack}
This research project was funded by the National Research Center for Applied Cybersecurity ATHENE, the EU project EXPLAIN, funded by the German Federal Ministry of Education and Research (BMBF) (grant 01—S22030D), as well as the Honda Research Institute Europe (HRI-EU). 
Furthermore, authors received support from the Eindhoven Artificial Intelligence Systems Institute as well as via the DEPTH group CAUSE of the Hessian Center for AI (hessian.ai).
It also benefited from the BMBF AI lighthouse project, the Hessian research priority programme LOEWE within the project WhiteBox, the HMWK cluster projects ``The Third Wave of AI'' and ``The Adaptive Mind'', the German Center for Artificial Intelligence (DFKI) project ``SAINT''.
\end{ack}


\bibliography{asn}
\twocolumn[\null]


\appendix
This technical appendix to Answer Set Networks provides additional details on the message passing, experimental setup, ASN programs with their associated reasoning graphs, and other supplementary information.

\section{Additional Details on Message-Passing}\label{app:sec:formal_asn}
Here, we want to address the message-passing once again, completing the node values update for the aggregate functions of $\min$ and $\max$. 
These were not discussed in the main paper to keep the formal definitions to the tasks considered in experiments.
With other words, the reader can skip all the descriptions and focus entirely on the Eq.~\eqref{eq:rg_complete_update_rule}.

Given a starting configuration of nodes that are considered true, we propagate truth values through the graph with the goal of computing the (stable) models of the program. 
Thus, we associate with every node $v\in V$ a boolean \textit{node value} $h_{v,t}\in\{0,1\}$ at a time step $t\in\mathbb{N}$. 
Except for $\top$, all nodes are initially set to false (0) at $t$ = 0.

Aggregate nodes $v\in\mathcal{V},\tau_v\in\mathcal{R}_\mathrm{aggr}$ additionally have terms $\beta_{v,l},\beta_{v,r}\in\mathbb{N}$ and relation operators $\diamond_{v,l},\diamond_{v,r}\in\{=,\neq,<,>,\le,\ge\}$ linked to them, representing the guards of the corresponding aggregate.
Furthermore, each edge $e\in\mathcal{E}$ has a \textit{weight}
\begin{equation}
    w_e\in
    \begin{cases}
        \{-1,1\}&\textrm{if }e\in\mathcal{E}_\mathrm{logical},\\
        \mathbb{Z} & \textrm{if }e\in\mathcal{E}_\mathrm{aggr}.
    \end{cases} 
\end{equation}
We can then define the node values for the next time step recursively as $\mathrm{h}_{v,t+1} =$
\begin{equation}
    \resizebox{.95\columnwidth}{!}{$
    \begin{cases}
        \mathrlap{\displaystyle \bigwedge_{\substack{e=(u,v)\in\mathcal{E},\\w_e=1}} h_{u,t} \land \bigwedge_{\substack{e=(u,v)\in\mathcal{E},\\w_e=-1}} \neg h_{u,t}} & \text{if } \tau_v=A_\land, \vspace{4pt}\\
        \mathrlap{\displaystyle \bigvee_{\substack{e=(u,v)\in\mathcal{E},\\w_e=1}} h_{u,t} \lor \bigvee_{\substack{e=(u,v)\in\mathcal{E},\\w_e=-1}} \neg h_{u,t}} & \text{if } \tau_v=A_\lor, \vspace{4pt}\\
        \beta_{v,l}\diamond_{v,l} |\mathcal{E}^{\mathrm{in},\top}_t(v)| \diamond_{v,r}\beta_{v,r} & \text{if } \tau_v=A_\mathrm{count}, \vspace{4pt}\\
        \beta_{v,l}\diamond_{v,l} \displaystyle\sum_{e\in\mathcal{E}^{\mathrm{in},\top}_t(v)} w(e) \diamond_{v,r}\beta_{v,r}  & \text{if } \tau_v=A_\mathrm{sum}, \vspace{4pt}\\
        \beta_{v,l}\diamond_{v,l} \min^{\preceq}(\mathcal{W}^{\mathrm{in},\top}_t(v)\cup\{\mathrm{\#sup}\}) \diamond_{v,r}\beta_{v,r} & \text{if } \tau_v=A_\mathrm{min}, \vspace{4pt}\\
        \beta_{v,l}\diamond_{v,l} \max^{\preceq}(\mathcal{W}^{\mathrm{in},\top}_t(v)\cup\{\mathrm{\#inf}\}) \diamond_{v,r}\beta_{v,r} & \text{if } \tau_v=A_\mathrm{max}.
    \end{cases}
    $}
    \label{eq:rg_complete_update_rule}
\end{equation}
where $\mathcal{E}^{\mathrm{in},\top}_t(v):=\{e|e=(u,v)\in\mathcal{E},h_{u,t}=1\}$ is the set of incoming edges of $v\in\mathcal{V}$ whose source nodes are 1 (true) at time step $t\in\mathbb{N}$ and $\mathcal{W}^{\mathrm{in},\top}_t(v):=\{w_e|e\in\mathcal{E}^{\mathrm{in},\top}_t(v)\}$ the set of corresponding edge weights.\\
\indent Updates at the time step $t+1$ can be understood in the following way. 
Logical nodes $v\in\mathcal{V}_\mathrm{logical}$ take on the value of the conjunction or disjunction of their incoming neighbors' values, where the edge weight determines whether the corresponding node value is included positively or negatively.
We therefore call an edge $e\in\mathcal{E}_\mathrm{logical}$ \emph{positive} if $w_e=1$ and \emph{negative} otherwise.
For aggregate nodes, $v\in\mathcal{V}_\mathrm{aggr}$, the edge weights of incoming edges signify the values to be aggregated, but are only included if the source node is considered true. 
The node value of an aggregate node is then determined by checking the guards of the aggregated value. For $\mathrm{\#min}$ and $\mathrm{\#max}$ aggregates, we include the default value for empty sets.
I.e., $\mathrm{\#sup}$ and  $\mathrm{\#inf}$ for an infinite $T$ and swapping both otherwise.

\section{How big are Reasoning Graphs?}
Tab.~\ref{tab:graph_size} lists the number of nodes and respectively edges for each experiment in the paper. 
\begin{table}[t]
    \centering
    \begin{tabular}{c|cc}
    \toprule
        Experiment & Nodes & Edges \\
        \midrule
         LLM Fine-Tuning & 162 & 106 \\
         ProMis & 122 & 56 \\
         MNIST-Add T1 & 495 & 354\\
         MNIST-Add T2 & 3277 & 4076\\
         MNIST-Add T3 & 30367& 50102\\
         \bottomrule
    \end{tabular}
    \caption{Nodes and Edges in the RG}
    \label{tab:graph_size}
\end{table}
We can observe that with increasing task complexity, the number of nodes and edges increases as well.
As discussed in the conclusion, the growing number of NPPs results in more choice definizations to be considered.
The definizations make up the node and edge dimensionality, and as for MNIST-Add T3 it can be memory intensive.
This concern should be addressed in future work.

\section{Experimental Details}\label{app:sec:exper_dits}
\textbf{Training runs}
We use multiple seeds to evaluate training time and performance for all experiments.
We use 3 seeds for LLM fine-tuning and ProMis, and 5 seeds for MNIST Addition.

\subsection{LLM Fine-Tuning}

\textbf{Training Setup -- } 
For training, we ask the following query: ``What is the relation between $<$person1$>$ and $<$person2$>$? The answer is:''. 
We then predict the next token and obtain the logits for the four relation classes: mother, father, daughter and son.
There are two directions we are interested in:
\begin{itemize}
    \item \textbf{parent-child}: ``What is the relation between $<$celeb\_parent\_name$>$ and $<$celebrity\_name$>$? The answer is:''$\Rightarrow$ mother or father
    \item \textbf{child-parent}: ``What is the relation between $<$celebrity\_name$>$ and $<$celeb\_parent\_name$>$? The answer is:'' $\Rightarrow$ daughter or son
\end{itemize}

\cite{reversalcurse} provided the dataset consisting of 797 unique celebrity names, 81 of which have one parent in the dataset and 716 with two parents in the dataset. 
This results in $81+716*2=1513$ parent-child and additional reversed, 1513 child-parent relations in total. 

For training, we provide the model with both directions but will never give them both directions for a single celebrity. 
If the model sees Tom Cruise and Mary Lee Pfeiffer in that order, it will never see Mary Lee Pfeiffer and Tom Cruise in that order.

\textbf{Data splits  -- }We split up the unique celebrity names into 70\% for training, 20\% for testing, and 10\% for the validation dataset.
From 557 unique celebrity names drawn at random for training, we obtain 1054 parent-child combinations as we have on average slightly less than two parents. 
Next, we randomly split them into 50\% child-parent and 50\% parent-child pairs.

Similarly, the test and validation datasets are built, but we keep them separate as shown in Fig.~5 (a) (p. 6). 

\textbf{Logic Program --}
While the baseline will only get a training signal from the data points in one direction, ASN will infer the relation from the other direction using the program in Listing~\ref{lst:app:aft}. 
To infer if the reverse relation is mother or father for parent-child relations or daughter or son for child-parent relations, the sex of the second person is required.
For this, the program encodes a call to the LLM that returns the probability for sex.
This way, we don't need to provide the data ourselves but can leverage the knowledge stored in the LLM itself.

\subsection{ProMis}\label{app:subsec:promis}
\begin{table}[t]
    \centering
    \resizebox{\columnwidth}{!}{
        \begin{tabular}{l|rrrrrrrrr}
        \toprule
         BS&25&50&250&500&2.5k&5k&25k&50k&250k\\
         Time&\ctimeG{7}{21}&\ctimeG{3}{41}&\ctimeG{1}{2}&\ctimeG{0}{42}&\ctimeG{0}{27}&\ctimeG{0}{24}&\ctimeG{0}{22}&\ctimeG{0}{22}&\textbf{\ctimeG{0}{21}}\\      
         \bottomrule    
        \end{tabular}
    }
    \caption{\textbf{Batching ProMis with ASN:} 
    Runtime comparison to compute a 500\textsuperscript{2} grid with different batch sizes in ASN. 
    Increasing the batch steadily reduces the computation time.}
    \label{tab:promis-batching}
\end{table}
In ASN, ProbLog, and SLASH, we must solve 6500\textsuperscript{2} programs corresponding to all coordinate query pairs to render Paris with each point corresponding to 2m\textsuperscript{2}.
Due to the graph-based nature, ASN can batch multiple queries together for parallel computation on the GPU, as described at the end of Sec. 2.2 (Neural Compilation) 
Consequently, we first want to evaluate the effect of batching in ASN by comparing increasing batch sizes. Table \ref{tab:promis-batching} shows the impact of different batch sizes on the time it takes to compute a grid with a resolution of 500\textsuperscript{2}. 
We observe that batching is an effective strategy to speed up the computation. After reaching a certain batch size, further gains are only marginal. For the ProMis Paris experiment, we used a high batch size of 500\textsuperscript{2}. 

While in ASN, we can use batching to speed up the compute time, we can group multiple coordinates together in one ProbLog program as in \cite{kohaut2023md}. 
We found that using 16 coordinates per program resulted in the fastest time and used it to compare with ASN. For SLASH we can use multiple CPU's in parallel and found 20 CPUs to be the fastest.

\subsection{MNIST-Addition}\label{app:subsec:mnist-add}
\begin{table}[t]
    \begin{threeparttable}
    \resizebox{\columnwidth}{!}{%
        \begin{tabular}{l|rlrlrl}
        \toprule
        & \multicolumn{6}{c}{Total Time (\#Epochs)}\\
        & \multicolumn{2}{c}{T1}& \multicolumn{2}{c}{T2}& \multicolumn{2}{c}{T3} \\
        Batch Size & time & e & time & e & time & e \\
        \midrule
        64 & 
        \ctimeD{1}{20} &\cellcolor{lastcolor}(2)&
        \ctimeE{6}{51}&\cellcolor{lastcolor}(2)&
        \ctimeF{1}{1}{13}&\cellcolor{lastcolor}(2)\\
        128 & 
        \ctimeD{0}{45}&\cellcolor{lastcolor}(2)&
        \ctimeE{3}{36}&\cellcolor{lastcolor}(2)&
        \ctimeF{0}{47}{5}&\cellcolor{lastcolor}(3)\\
        256 & 
        \ctimeD{0}{26}&\cellcolor{lastcolor}(2)&
        \ctimeE{2}{46}&\cellcolor{lastcolor}(3)&
        \ctimeF{0}{24}{10}&\cellcolor{lastcolor}(3)\\
        512 & 
        \ctimeD{0}{24}&\cellcolor{lastcolor}(3)&
        \ctimeE{1}{59}&\cellcolor{lastcolor}(4)&
        \ctimeF{0}{21}{10}&\cellcolor{lastcolor}(5)\\
        1024 & 
        \textbf{\ctimeD{0}{22}}&\cellcolor{lastcolor}(4)& 
        \ctimeE{1}{35}&\cellcolor{lastcolor}(6)&
        \ctimeF{0}{17}{11}&\cellcolor{lastcolor}(8)\\
        2048 & 
        \ctimeD{0}{31}&\cellcolor{lastcolor}(7)& 
        \textbf{\ctimeE{1}{32}}&\cellcolor{lastcolor}(10)&
        \ctimeF{0}{16}{54}&\cellcolor{lastcolor}(15)\\
        4096 & 
        \ctimeD{0}{46}&\cellcolor{lastcolor}(12)& 
        \ctimeE{1}{53}&\cellcolor{lastcolor}(19)&
        \textbf{\ctimeF{0}{15}{51}}&\cellcolor{lastcolor}(27)\\
        8192 & 
        \ctimeD{1}{23}&\cellcolor{lastcolor}(23)& 
        \ctimeE{2}{53}&\cellcolor{lastcolor}(38)&
        \ctimeF{0}{15}{46}&\cellcolor{lastcolor}(50*)\\
        15000 & 
        \ctimeD{2}{50}&\cellcolor{lastcolor}(49)& 
        \ctimeE{3}{8}&\cellcolor{lastcolor}(48)&
        \ctimeF{0}{9}{3}&\cellcolor{lastcolor}(50*)\\
        \bottomrule
        \end{tabular}
        }
    \begin{tablenotes}\footnotesize
        \item[*] Did not reach the accuracy threshold within the maximum\\ number of epochs.
    \end{tablenotes}
    \end{threeparttable}
    \caption{\textbf{Batch size trade-off for MNIST-addition:} Models were trained for a maximum of fifty epochs to achieve a competitive accuracy of $\ge98\%$. Results for 30k are not shown as they did not converge in 50 epochs, and the dataset size for T2 and T3 is 20k and 15k, respectively.}
    \label{tab:mnist-acc-tradeoff}
\end{table}
\textbf{Computation of Batch Size -- } Having seen the scaling abilities of ASN for ProMis, it is beneficial to address how a chosen batch size influences both the resulting accuracy and the number of epochs until the convergence point. 
Tab.~\ref{tab:mnist-acc-tradeoff} shows the total training time (average time per epoch multiplied by the average number of trained epochs to reach a threshold of 98\%).
The numbers in parentheses after the reported times indicate the number of epochs needed.
Note that for T1 and a batch size of 30k, all runs did not meet the accuracy criteria within the maximum number of epochs. Furthermore, for T2 and T3, the data sets contain a maximum of 20k and 15k training points, respectively, which is why the largest batch size we report is 15k.

Following the results, we see a non-obvious optimum for every complexity level. The bigger the batch size, the faster it is to train for one epoch. As neural network training is involved in this experiment, we cannot simply choose the largest batch size to lower compute time, as in ProMis: The higher the batch size, the longer it takes the network to converge. 
Consequentially, for every task, it is necessary to tune the batch size to achieve the maximum speed-up without compromising the overall accuracy.

\textbf{Computation Speed -- } After seeing the influence of choosing the batch size, let us look at what happens if we compare the average time per epoch following the setting proposed in ~\citep{SAME}.
Therefore, we use DeepProbLog \cite{DeepProbLog}, SLASH, and SAME as baselines. 
Having reimplemented SLASH with ASN, we thus have a fair comparison.
Tab. 1 lists the results showing ASN scalability advantages.
It tightly binds the end-to-end computations for WMC~\cite{introduction_to_WMC} in a vectorized way so that even SAME dynamically pruning the potential solutions is not keeping pace.
Thereby, it is noticeable that the bigger the solution space of the task at hand, the greater the speed-up ASN allows. 
As a result, we argue that ASN paves the way for vastly complex and, until now, only cumbersome solvable tasks in NeSy AI.
\begin{figure*}[t!]
    \centering
    \resizebox{1.0\linewidth}{!}{
        \includegraphics[]{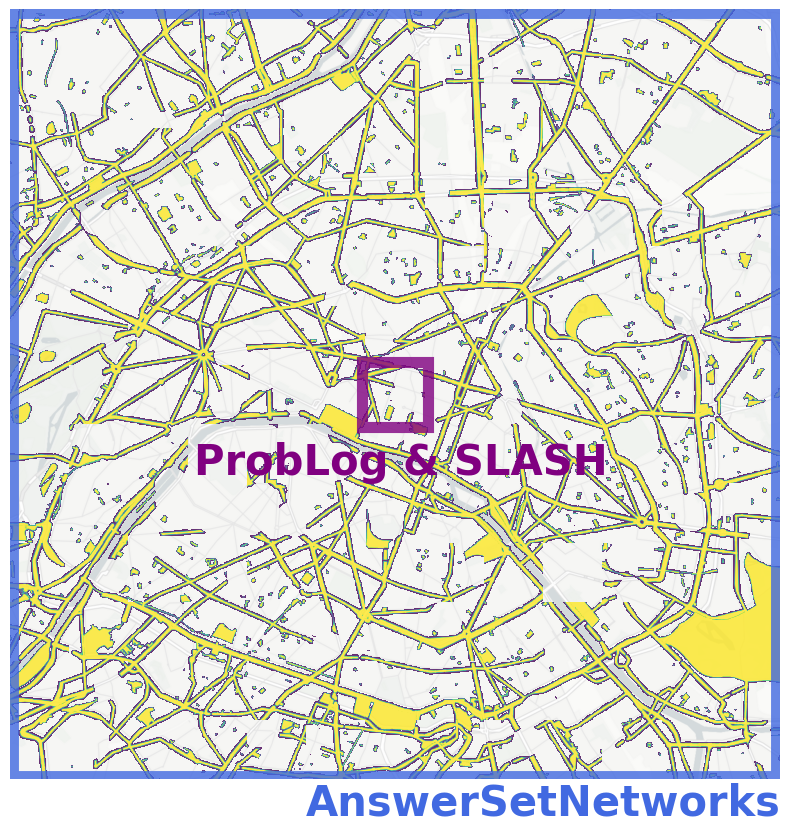}
    }
    \caption{\textbf{Visualization ProMis Paris in full size}}
    \label{fig:promis-big}
\end{figure*}

\section{Programs}\label{apx:sec:programs}
Here, the interested reader will find the ASP programs which we compiled for our ablation studies on LLM fine-tuning trough probabilistic logic, ProMis over Paris and MNIST-addition.
The RGs for the programs here are automatically generated with \textit{Graphviz} package and are colored differently than in the main paper.\\

\begin{figure*}
    \centering
    \resizebox{1.0\linewidth}{!}{
        \includegraphics[]{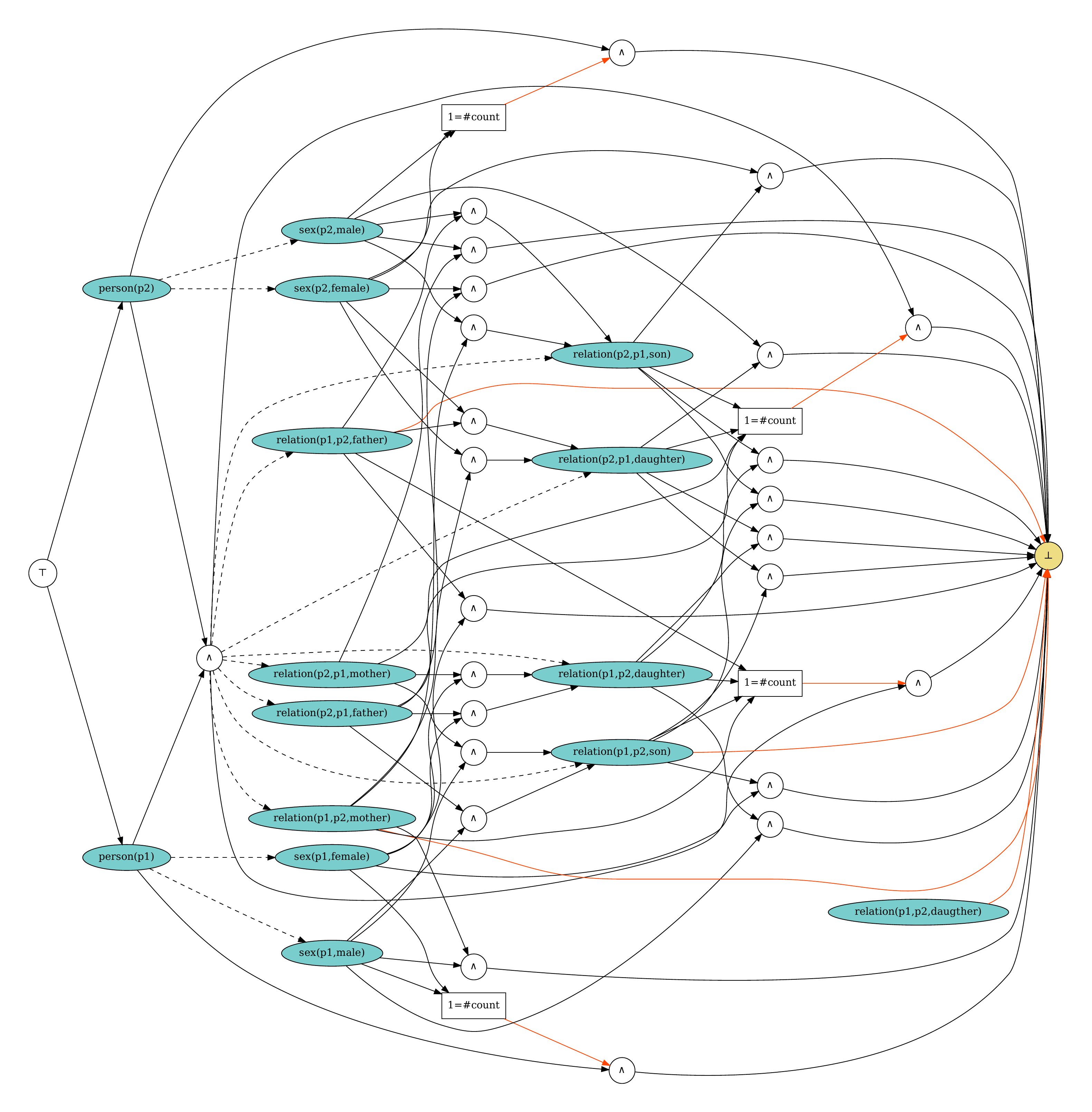}
    }
    \caption{
        \textbf{\textit{Reasoning Graph} for LLM Fine-Tuning with ASN:}
        We query for all four possible relationship types among two persons and pick the one with the highest probability.
    }
    \label{img:app:rg_aft}
\end{figure*}

\begin{figure*}
    \centering
    \resizebox{1.0\linewidth}{!}{
        \includegraphics[]{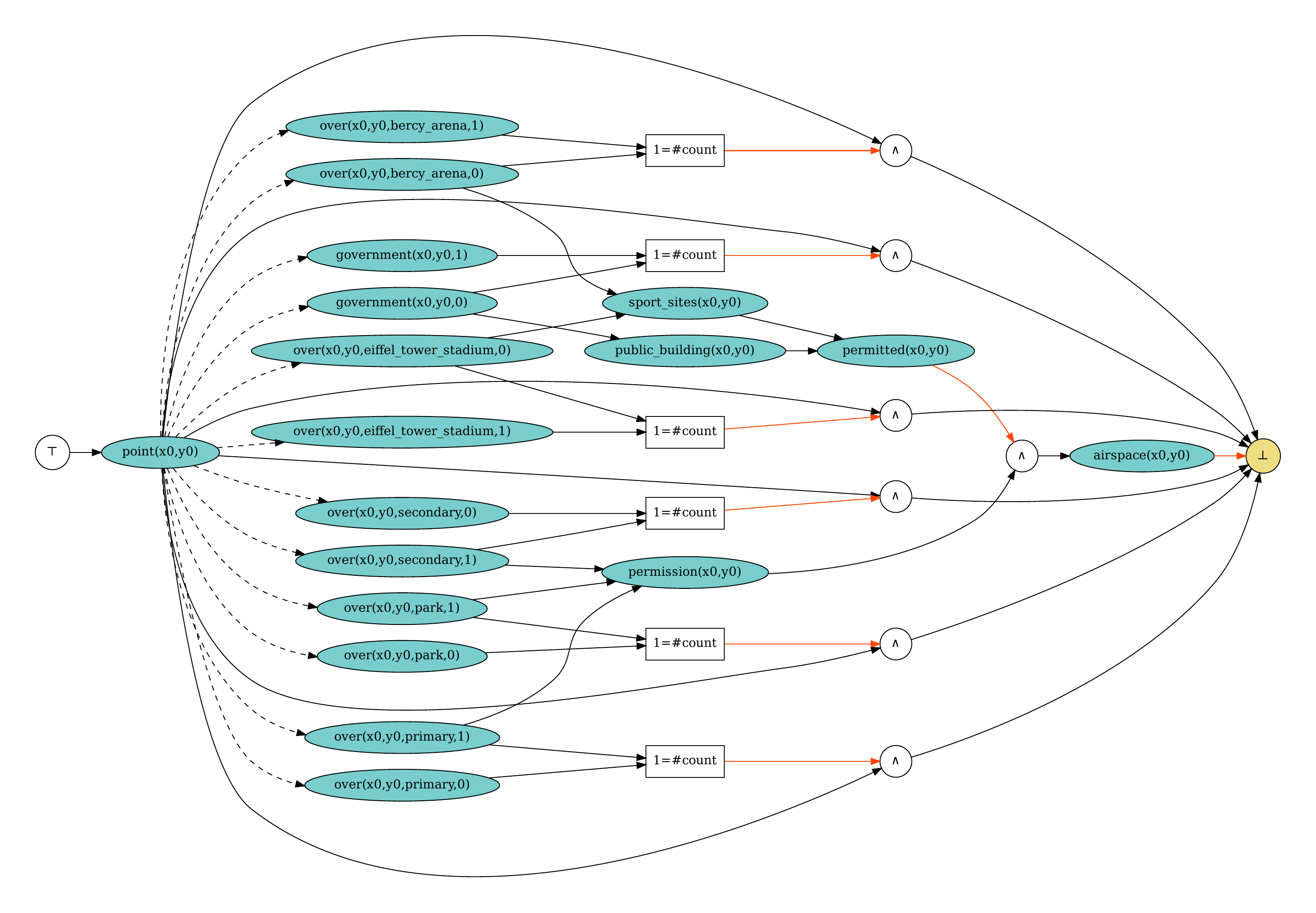}
    }
    \caption{
        \textbf{\textit{Reasoning Graph} for ProMis on Paris:}
        Only one grid point is listed as query for the RG to become easily displayable. 
    }
    \label{img:app:rg_promis_paris}
\end{figure*}

\begin{figure*}
    \centering
    \resizebox{1.0\linewidth}{!}{
        \includegraphics[]{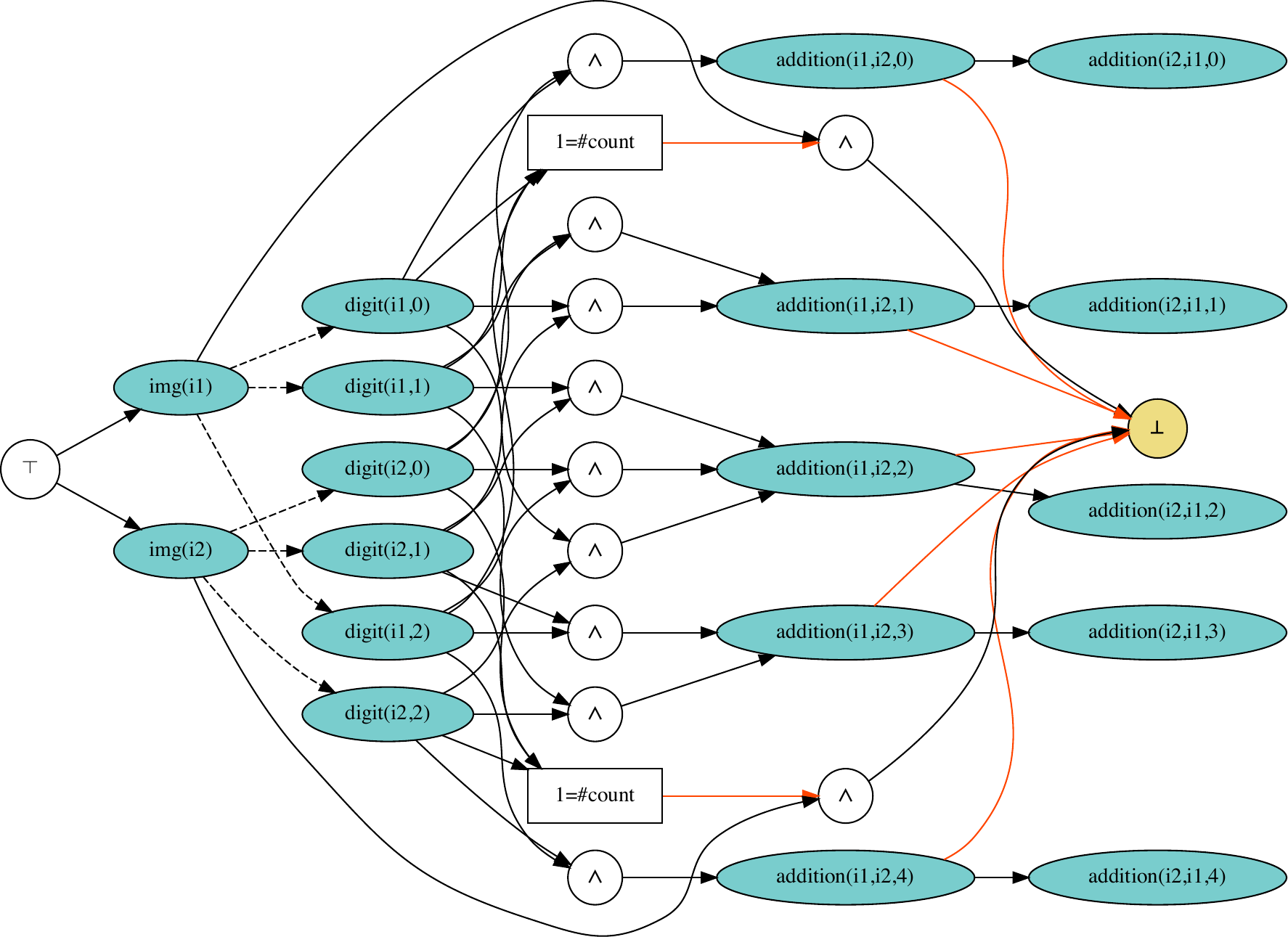}
    }
    \caption{
        \textbf{\textit{Reasoning Graph} for MNIST-Addition:}
        The number of classes was restricted to $[0,1,2]$ for the RG to become easily displayable.  
    }
    \label{img:app:rg_mnist_addition}
\end{figure*}

\noindent\textbf{LLM Fine-Tuning with ASN}
Listing~\ref{lst:app:aft} shows the SLASH program used for the task, and Fig.~\ref{img:app:rg_aft} outlines its RG.\\ 
\noindent\textbf{ProMis}
Listing~\ref{lst:app:promis} shows the simplified version of the source code, and Fig.~\ref{img:app:rg_promis_paris} offers its RG.\\
\noindent\textbf{MNIST-Addition}
Listing~\ref{lst:app:mnist-addition} lists the source code for the task, while Fig.~\ref{img:app:rg_mnist_addition} shows the according RG.

\begin{listing}
    \caption{LLM fine-tuning through probabilistic logic to alleviate the ``Reversal Curse''}
    \label{lst:app:aft}
    \begin{lstlisting}
% We use the same LLM playing the role of both NPPs. I.e., two forward passes to obtain the probabilities.
person(p1). person(p2).

% NPP to determine the relation of two different persons
#npp(relation(X1,X2),[mother,father,daughter,son]) :- person(X1), person(X2), X1 != X2.

% NPP to determine the sex of a person
#npp(sex(X),[male,female]) :- person(X).

% Exclude sexes being open.
:- relation(X1,X2,mother), sex(X1,male).
:- relation(X1,X2,father), sex(X1,female).
:- relation(X1,X2,son), sex(X1,female).
:- relation(X1,X2,daughter), sex(X1,male).

% Constraining relations to be acyclic.
:- relation(X1,X2,daughter), relation(X2,X1,son).
:- relation(X1,X2,son), relation(X2,X1,son).
:- relation(X1,X2,daughter), relation(X2,X1,daughter).

% Constraining possible relations
% C1: If the child is male, than it is a son.
relation(X2,X1,son) :- relation(X1,X2,mother), sex(X2,male).
relation(X2,X1,son) :- relation(X1,X2,father), sex(X2,male).

% C2: If the child is female than it is a daughter
relation(X2,X1,daughter) :- relation(X1,X2,mother), sex(X2,female).
relation(X2,X1,daughter) :- relation(X1,X2,father), sex(X2,female). 

% Query
:- not relation(p1,p2,daugther).
:- not relation(p1,p2,son).
:- not relation(p1,p2,mother).
:- not relation(p1,p2,father).
    \end{lstlisting}
\end{listing}

\begin{listing}
    \caption{ProMis over Paris ASP-program. Only one query included for illustration.}
    \label{lst:app:promis} 
    \begin{lstlisting}
% Single pixel defined by latitude and longitude as typle (x,y)
point(x0,y0). 

% Batch of probabilistic facts       
#npp(over(X,Y,park), [1,0]):- point(X,Y).
#npp(over(X,Y,primary), [1,0]):- point(X,Y).
#npp(over(X,Y,eiffel_tower_stadium), [1,0]):- point(X,Y).
#npp(over(X,Y,bercy_arena), [1,0]):- point(X,Y).
#npp(over(X,Y,secondary), [1,0]):- point(X,Y).
#npp(embassy(X,Y), [1,0]):- point(X,Y).
#npp(government(X,Y), [1,0]):- point(X,Y).

% Its okay to go over park areas
permission(X,Y) :- over(X,Y,park,1).

% Its okay to fly over major roads
permission(X,Y) :- over(X,Y,primary,1).
permission(X,Y) :- over(X,Y,secondary,1).

% Define sport sites
sport_sites(X,Y) :- over(X,Y,eiffel_tower_stadium,0).
sport_sites(X,Y) :- over(X,Y,bercy_arena,0).

% Define public buildings
public_building(X,Y) :- embassy(X,Y,0).
public_building(X,Y) :- government(X,Y,0).

% It is not allowed to fly over sport sites and public buildings
permitted(X,Y) :- sport_sites(X,Y).
permitted(X,Y) :- public_building(X,Y).

% We are only permitted to fly over the permitted areas which are not restricted otherwise
airspace(X,Y) :- permission(X,Y), not permitted(X,Y).
                       
% Query
:- not airspace(x0,y0).
    \end{lstlisting}%
\end{listing}

\begin{listing}
    \caption[MNIST Addition in ASN.]{MNIST Addition ASP-program. Every image is either a zero, a one or a two.
    Queries are just an example of potential outcomes of addition.}
    \label{lst:app:mnist-addition} 
    \begin{lstlisting}
% Both images are facts of the program
img(i1). img(i2).

% NPP-rule 
#npp(digit(X), [0,1,2]) :- img(X).

% Addition
addition(A,B,N1+N2) :- digit(A,N1), digit(B,N2), A<B.

% Exclude symmetries
addition(A,B,N) :- addition(A,B,N), A<B.

% Queries
:- not addition(i1,i2,0).
:- not addition(i1,i2,1).
:- not addition(i1,i2,2).
:- not addition(i1,i2,3).
:- not addition(i1,i2,4).
    \end{lstlisting}%
\end{listing}

\clearpage

\end{document}